\documentclass[10pt,journal,compsoc]{IEEEtran}
\ifCLASSOPTIONcompsoc
  \usepackage[nocompress]{cite}
\else
  \usepackage{cite}
\fi
\ifCLASSINFOpdf
\else
\fi
\usepackage{epsfig}
\usepackage{graphicx}
\usepackage{amsmath}
\usepackage{amssymb}
\usepackage{adjustbox}
\usepackage{booktabs}
\usepackage{gensymb}
\usepackage[hidelinks]{hyperref}
\usepackage{color, colortbl}
\usepackage{multirow}
\usepackage{capt-of}
\usepackage{dblfloatfix}

\usepackage{pifont}

\definecolor{aliceblue}{rgb}{0.94, 0.97, 1.0}
\def\degree{${}^{\circ}$}

\usepackage[dvipsnames]{xcolor}

\newcommand{\wang}[1]{\textcolor{black}{#1}}
\newcommand{\jb}[1]{\textcolor{black}{#1}}
\newcommand{\ang}[1]{\textcolor{black}{#1}}

\newcommand{\etal}{\textit{et al}.}
\newcommand{\ie}{\textit{i}.\textit{e}.}
\newcommand{\eg}{\textit{e}.\textit{g}.}
\newcommand{\etc}{\textit{etc}.}
\newcommand{\vs}{\textit{vs.}}
\newcommand{\cc}{\cellcolor{aliceblue}}
\newcommand{\xmark}{\ding{55}}

\begin{document}
%
\title{Self-supervised Video Representation Learning by Uncovering Spatio-temporal Statistics}
%
%
%
%

\author{Jiangliu~Wang$^*$,~Jianbo~Jiao$^*$,~\IEEEmembership{Member,~IEEE},~Linchao~Bao,~Shengfeng~He,~\IEEEmembership{Senior Member,~IEEE}, \\ Wei Liu,~\IEEEmembership{Senior Member,~IEEE} and
	~Yun-hui~Liu,~\IEEEmembership{Fellow,~IEEE}
	
\IEEEcompsocitemizethanks{
		\IEEEcompsocthanksitem J. Wang and Y. Liu are with the Chinese University of Hong Kong (CUHK), the CUHK T Stone Robotics Institute, and the Hong Kong Centre for Logistics Robotics. Email: \{jiangliuwang@link.cuhk.edu.hk, yhliu@cuhk.edu.hk\}.
		\IEEEcompsocthanksitem J. Jiao is with the Department of Engineering Science, 
		University of Oxford. Email: jianbo@robots.ox.ac.uk.
		\IEEEcompsocthanksitem L. Bao and W. Liu are with Tencent AI Lab. Email: \{linchaobao@gmail.com, wl2223@columbia.edu\}
		\IEEEcompsocthanksitem S. He is with the School of Computer Science and Engineering, 
		South China University of Technology. Email: shengfenghe7@gmail.com.
		\IEEEcompsocthanksitem J. Wang and J. Jiao contributed equally to this work. L. Bao and Y. Liu are the corresponding authors. 
	}
}

\markboth{IEEE Transactions on Pattern Analysis and Machine Intelligence,~Vol.~xx, No.~x, January~2021}%
{Shell \MakeLowercase{\textit{et al.}}: Bare Demo of IEEEtran.cls for Computer Society Journals}
%



\IEEEtitleabstractindextext{%
\begin{abstract}
This paper proposes a novel pretext task to address the self-supervised video representation learning problem. 
Specifically, given an unlabeled video clip, we compute a series of spatio-temporal statistical summaries, such as the spatial location and dominant direction of the largest motion, the spatial location and dominant color of the largest color diversity along the temporal axis, \etc~
Then a neural network is built and trained to yield the statistical summaries given the video frames as inputs.
In order to alleviate the learning difficulty, we employ several spatial partitioning patterns to encode rough spatial locations instead of exact spatial Cartesian coordinates. 
Our approach is inspired by the observation that human visual system is sensitive to rapidly changing contents in the visual field, and only needs impressions about rough spatial locations to understand the visual contents. 
To validate the effectiveness of the proposed approach, we conduct extensive experiments with four 3D backbone networks, \ie, C3D, 3D-ResNet, R(2+1)D and S3D-G. 
The results show that our approach outperforms the existing approaches across these backbone networks on four downstream video analysis tasks including action recognition, video retrieval, dynamic scene recognition, and action similarity labeling.
The source code is publicly available at: \url{https://github.com/laura-wang/video_repres_sts}. 
\end{abstract}

}

\maketitle

\IEEEdisplaynontitleabstractindextext

%
\IEEEpeerreviewmaketitle

\IEEEraisesectionheading{\section{Introduction}\label{sec:introduction}}

\IEEEPARstart{P}{owerful} video 
\ang{representation serves as a foundation}
for solving many video content analysis and understanding tasks, such as action recognition~\cite{carreira2017quo, r2plus1d_cvpr18}, video retrieval~\cite{liu2019use, miech2019howto100m}, video captioning~\cite{wang2018reconstruction, wang2018bidirectional}, \etc~Various network architectures~\cite{simonyan2014two, carreira2017quo, tran2018closer} are designed and trained with massive human-annotated video data to learn video representation for individual tasks. While great progresses have been made, supervised video representation learning is impeded by two major obstacles: (1) Annotation of video data is labour-intensive and expensive, restricting supervised learning to relish a large quantity of free video resources on the Internet. (2) Representation learned from labeled video data lacks generality and robustness, \eg, video features learned for action recognition do not well to video retrieval task~\cite{buchler2018improving, luo2020video}.

To tackle the aforementioned challenges, multiple approaches~\cite{misra2016shuffle, lee2017unsupervised, kim2018self, xu2019self} have emerged to learn more generic and robust video representation in a self-supervised manner. Neural networks are first pre-trained with unlabeled videos using 
\emph{pretext tasks}, where supervision signals are derived from input data without human annotations. Then the learned representation can be employed as weight initialization for training models or be directly used as features in succeeding \emph{downstream tasks}. 

\begin{figure}[t]
	\centering
	\includegraphics[width=\columnwidth]{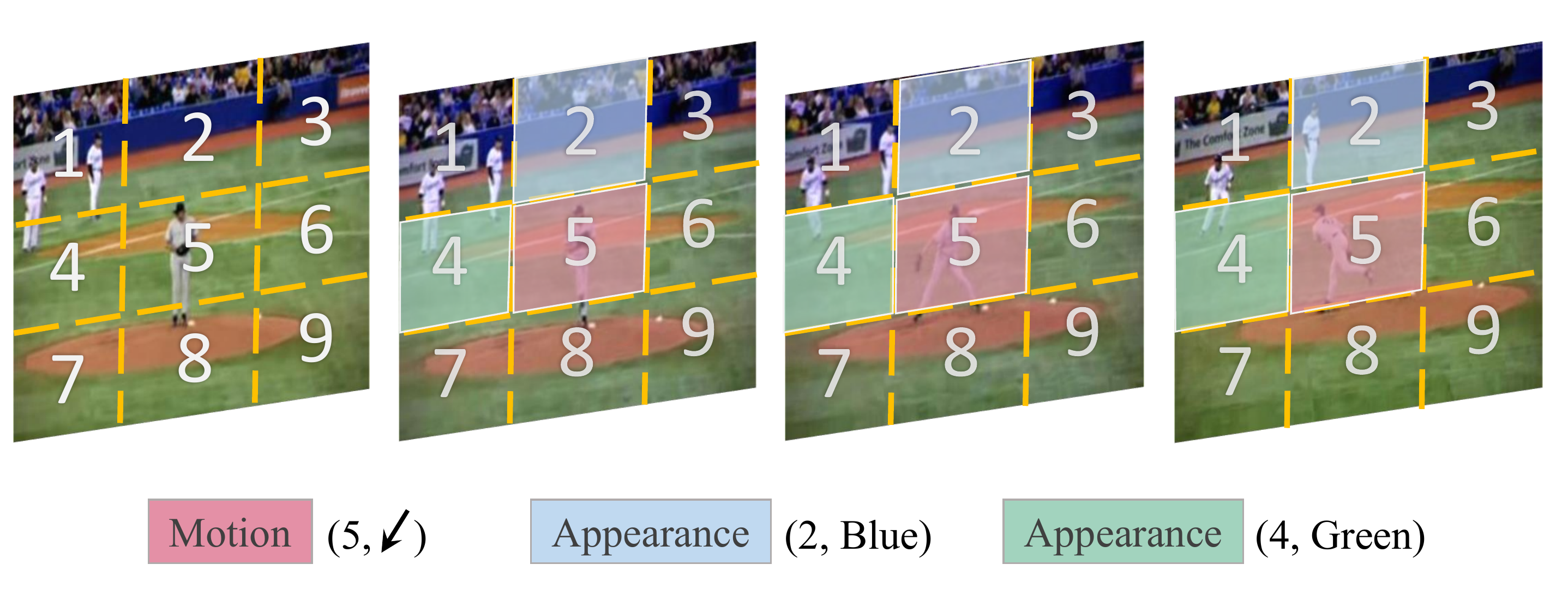}
	\caption{Main idea of the proposed approach. \wang{Given a video sequence, we design a pretext task to uncover the summaries derived from spatio-temporal statistics for self-supervised video representation learning.} 
	Specifically, each video frame is first divided into several spatial regions using different partitioning patterns like the grid shown in the figure. Then the derived statistical labels, such as \emph{the region with the largest motion and its direction} (the red patch), \emph{the most diverged region in appearance and its dominant color} (the blue patch), and \emph{the most stable region in appearance and its dominant color} (the green patch), are employed as supervision signals to guide the representation learning.}
	\label{fig:teas}
\end{figure}
 
\wang{Among the existing self-supervised video representation learning methods, video order verification/prediction~\cite{misra2016shuffle, fernando2017self, lee2017unsupervised, kim2018self,xu2019self} is one of the most popular pretext tasks. It randomly shuffles video frames and asks a neural network to predict whether the video is perturbed or to rearrange the frames in a correct chronological order. 
By utilizing the intrinsic temporal characteristics of videos, these pretext tasks have been shown useful for learning high-level semantic features. 
Other approaches include flow fields prediction~\cite{gan2018geometry}, future frame prediction~\cite{vondrick2016generating, lotter2016deep, mathieu2016deep}, dense predictive coding~\cite{han2019video}, \etc~}Although promising results have been achieved, 
the dense prediction pretext tasks
may lead to redundant feature learning towards solving the pretext task itself, instead of learning generic representative features for downstream video analysis tasks. For example, predicting the future frame requires the network to precisely estimate each pixel in each frame in a video clip. This increases the learning difficulties and causes the network to waste a large portion of capacity on learning features that may be not transferable to high-level video analysis tasks.

\wang{In this paper, enlightened by the human visual system~\cite{giese2003cognitive},
we propose a novel pretext task to learn video representation by \jb{uncovering} spatio-temporal statistical summaries from unlabeled videos. For instance, given a video clip, the network is encouraged to identify the largest moving area with its corresponding motion direction, as well as the most rapidly changing region \jb{in appearance} with its dominant color. This idea is inspired by the cognitive study on human visual system \cite{giese2003cognitive}, in which the representation of motion is found to be based on a set of learned patterns.} These patterns are encoded as sequences of ``snapshots'' of body shapes by neurons in the \emph{form pathway}, and by sequences of complex optic flow patterns in the \emph{motion pathway}. In our work, these two pathways are defined as the appearance branch and motion branch, respectively. In addition, we define and extract several abstract statistical summaries accordingly, which is also inspired by the biological hierarchical perception mechanism~\cite{giese2003cognitive}.

We design several spatial partitioning patterns to encode each spatial location and its spatio-temporal statistics over multiple frames, and use the encoded vectors as supervision signals to train the neural network for spatio-temporal representation learning. The novel objectives are 
informative for the motion and appearance distributions in videos, \eg, the spatial locations of the most dominant motions and their directions, the most consistent and the most diverse colors over a certain temporal cube, \etc~An illustration of the main idea is shown in Fig.~\ref{fig:teas}, where a $3\times3$ grid pattern with motion and appearance statistics is shown for example. We conduct extensive experiments with 3D convolutional neural networks (CNNs) to validate the effectiveness of the proposed approach. The experimental results show that, compared with training from scratch, pre-training using our approach demonstrates a large performance gain for video action recognition problem (\eg, $56.0\%$ \vs~$77.8\%$ on UCF101 and $22.0\%$ \vs~$40.7\%$ on HMDB51).  By transferring the learned representation to other video tasks, such as video retrieval, dynamic scene recognition, and action similarity labeling, we further demonstrate the generality and robustness of the video representation learned by the proposed approach.

\wang{A preliminary version of this work was presented in~\cite{wang2019self}, where the basic idea of utilizing spatio-temporal statistical information for video representation learning is introduced. In this paper, we \jb{further} extend the previous work in five aspects: \jb{(1)} We provide a more detailed implementation of the proposed self-supervised learning approach. We extend the proposed method to four backbone networks, \ie, C3D with BN, 3D-ResNet, R(2+1)D, and S3D-G, on a large-scale dataset kinetics-400~\cite{kay2017kinetics}. 
(2) We conduct extensive ablation studies on the effectiveness of the pre-training dataset, the effectiveness of different bakcbone networks, and the correlation between pretext task and downstream task performances.
\jb{(3)} We investigate the effectiveness of different training targets, including 1D-label regression, 2D-label regression, and classification. \jb{(4)} A curriculum learning strategy is introduced to further improve the representation learning. \jb{(5)} We \jb{further} validate the proposed method on a new downstream task, video retrieval, to evaluate the \jb{generalizability of the learned representation.
}
}
         
To summarize, the main contributions of this work are four-fold: (1) We propose a novel pretext task for video representation learning by uncovering motion and appearance statistics without human annotated labels. 
(2) We introduce a curriculum learning strategy based on the proposed spatio-temporal statistics, which is also inspired by the human learning process: from simple samples to difficult samples. (3) Extensive ablation studies are conducted and analyzed to reveal several insightful findings for self-supervised learning, including the effectiveness of training data scale, network architectures, correlation between pretext task and downstream tasks,
and feature generalization, to name a few. (4) The proposed approach significantly outperforms previous approaches across all the studied network architectures in various video analysis tasks. Code and models are made publicly available online.

 

\section{Related Work}\label{sec:related}
In this section, we review related works, including self-supervised representation learning and its applications on downstream video analysis tasks. 
Please refer to a recent survey~\cite{jing2020self} for more details. 

\subsection{Self-supervised Representation Learning}

\wang{\textbf{Self-supervised Image Representation Learning.}
Self-supervised visual representation learning is first proposed and investigated in the image domain~\cite{doersch2015unsupervised}.
Various novel pretext tasks have been proposed to learn image representation from unlabeled image data, including predicting image context~\cite{doersch2015unsupervised}, re-ordering perturbed image patches \cite{noroozi2016unsupervised}, colorizing grayscale images \cite{zhang2016colorful}, inpainting missing regions \cite{pathak2016context}, counting virtual primitives \cite{noroozi2017representation}, classifying image rotations \cite{gidaris2018unsupervised}, predicting image labels obtained using a clustering algorithm \cite{caron2018deep}, \etc~There are also studies that try to learn image representation from unlabeled video data. Wang and Gupta \cite{wang2015unsupervised} proposed to derive supervision signals from unlabeled videos using traditional tracking algorithms. Instead Pathak \etal \cite{pathak2017learning} obtained supervision signals using conventional motion segmentation algorithms.}

\wang{Very recently, contrastive learning has yielded remarkable performance~\cite{chen2020simple,he2020momentum} and attracted wide attention in the self-supervised representation learning field. The intuition behind this approach is to conduct instance discrimination. It simultaneously minimizes the distances of positive pairs, and maximizes the distances of negative pairs in the latent space. Compared with the hand-crafted pretext tasks described above, this framework allows more flexibility of the design of self-supervised learning approaches. A line of works~\cite{oord2018representation,bachman2019learning,henaff2019data,wu2018unsupervised,tian2019contrastive} are thereby encouraged and introduced to resolve the most fundamental problem of contrastive learning: how to define the positive/negative pairs. 
Inspired by the empirical success, some works~\cite{purushwalkam2020demystifying,arora2019theoretical} have also made efforts to reveal the essential nature of contrastive learning in a more theoretical way. Readers are encouraged to view the self-supervised representation learning from both hand-crafted pretext tasks perspective and contrastive learning perspective, which provides a bigger picture.}

\wang{\textbf{Self-supervised Video Representation Learning.} Inspired by the success of self-supervised image representation learning, many works have emerged to learn transferable representation for video-related downstream tasks, such as action recognition, video retrieval, \etc~Intuitively, a large number of studies \cite{fernando2017self, lee2017unsupervised, misra2016shuffle, xu2019self, kim2018self} leveraged the distinct temporal information of videos and proposed to use frame sequence ordering as their pretext tasks. 
Wei \etal~\cite{wei2018learning} also proposed to predict whether the video clips are playing forwards or backwards.
B\"{u}chler \etal~\cite{buchler2018improving} further used deep reinforcement learning to design a sampling permutations policy. Gan \etal  \cite{gan2018geometry} proposed a geometry-guided network that forces the CNN to predict flow fields or disparity maps between two consecutive frames.} 

\wang{Although these work demonstrated the effectiveness of self-supervised representation learning with unlabeled videos and showed impressive performances when transferring the learned features to video recognition tasks, their approaches are only applicable to a CNN that accepts one or two frames as inputs and cannot be applied to network architectures that are suitable for spatio-temporal representation.
To address this problem, works~\cite{luo2020video, xu2019self, kim2018self} have been introduced to use 3D CNNs as backbone networks for spatio-temporal representation learning. Naturally, the 2D frame ordering pretext tasks are extended to 3D video clip ordering.
Some recent works~\cite{benaim2020speednet,epstein2020oops,jenni2020video,wang2020self} also demonstrated that predicting the video playback pace/speed is a simple-yet-effective pretext task. Inspired by the success of contrastive learning in the image domain, some works~\cite{han2019video,han2020memory}, also attempted to extend the concept of contrastive learning in the video domain.   
Another line of research should be mentioned is to leverage multi-modality sources, \eg, video-audio~\cite{owens2018audio, korbar2018cooperative} and video-text~\cite{miech2020end}, for self-supervised representation learning.
Note that in this paper, we focus on single modality, \ie, only consider learning representation in the video domain.}

\subsection{Representation Learning for Video Analysis Tasks}

Representation learning serves as a fundamental building block in tackling most video analysis tasks, such as complex action recognition \cite{hussein2019timeception}, action detection and localization \cite{chao2018rethinking, shou2017cdc, shou2016temporal}, video captioning \cite{wang2018reconstruction, wang2018bidirectional}, \etc~  
Two types of application modes are commonly adopted to evaluate the self-supervised video representation learning, either through transfer learning (as an initialization model) or feature learning (as a feature extractor).

Action recognition is one of the most widely used downstream video analysis tasks. At the beginning, researchers have developed various spatio-temporal descriptors for video representation to tackle this problem~\cite{laptev2005space, klaser2008spatio, wang2013action}. Promising results were achieved by improved dense trajectories (iDT) descriptors~\cite{wang2013action}, the best-performing hand-crated feature. Recently, extensive efforts have been focusing on the deep neural networks development due to the impressive success achieved by CNN. Tran \etal~\cite{tran2015learning} proposed C3D that extends the 2D kernels to 3D kernels to capture spatio-temporal video representation. Simonyan and Zisserman \cite{simonyan2014two} proposed a two-stream network that extracts spatio and temporal features on RGB and optical flow inputs, respectively. Stemmed from these two works, various network architectures are designed to learn video representation, including P3D~\cite{qiu2017learning}, I3D~\cite{carreira2017quo}, R(2+1)D~\cite{tran2018closer},~\etc~In this work, we consider to use three backbone networks, C3D\cite{tran2015learning}, 3D-ResNet\cite{tran2018closer} and R(2+1)D \cite{tran2018closer} to validate the proposed approach, following previous works~\cite{luo2020video, xu2019self}. Backbone networks pre-trained with the proposed spatio-temporal statistics will be used as weight initialization and fine-tuned on UCF101~\cite{soomro2012ucf101} and HMDB51~\cite{kuehne2011hmdb} datasets for the action recognition downstream task.

The other kind of evaluation mode is to use the pre-trained networks as feature extractors for the downstream video analysis tasks, such as video retrieval~\cite{xu2019self, luo2020video, misra2016shuffle}, dynamic scene recognition~\cite{gan2018geometry, derpanis2012dynamic}, \etc~Without fine-tuning, such a mode can directly evaluate the generality and robustness of the learned features. Performances of the self-supervised methods are compared with both competitive hand-crated video features, such as spatio-temporal interest points~\cite{laptev2005space},  HOG3D~\cite{klaser2008spatio}, slow feature analysis~\cite{theriault2013dynamic}, bags of spacetime energies~\cite{feichtenhofer2014bags}, \etc,~and other self-supervised learning methods.


\section{Our Proposed Approach}\label{sec:approach}

In this section, we first explain the high-level ideas and motivations for designing our novel pretext task with a simple illustration in Section~\ref{sec.concept}. Next, we formally define the computation of the spatio-temporal statistical labels from the motion aspect in Section~\ref{sec.motionstat} and appearance aspect in Section \ref{sec.rgbstat}.  A curriculum learning strategy is presented in Section \ref{sec:cl}. Finally, we summarize the whole learning framework with 3D CNNs in Section~\ref{sec.learnc3d}. 

\subsection{Motivation}\label{sec.concept}

Inspired by human visual system, we break the process of video contents understanding into several questions and encourage a CNN to answer them accordingly: (1) Where is the largest motion in a video? (2) What is the dominant direction of the largest motion? (3) Where is the largest color diversity and what is its dominant color? (4) Where is the smallest color diversity, \eg, the potential background of a scene, and what is its dominant color? 
The motivation behind these questions is that the human visual system \cite{giese2003cognitive} is sensitive to large motions and rapidly changing contents in the visual field, and only needs impressions about rough spatial locations to understand the visual contents.
We argue that a good pretext task should be able to capture necessary representation of video contents for downstream tasks, while does not waste model capacity on learning too detailed information that is not transferable to other downstream tasks.
~To this end, we design our pretext task as learning to answer the above questions with only rough spatio-temporal statistical summaries, \eg, for spatial coordinates we employ several spatial partitioning patterns to encode rough spatial locations instead of exact spatial Cartesian coordinates. 
In the following, we use a simple illustration to explain the basic idea.

\begin{figure}[t]
	\begin{center}
		\includegraphics[width=1\linewidth]{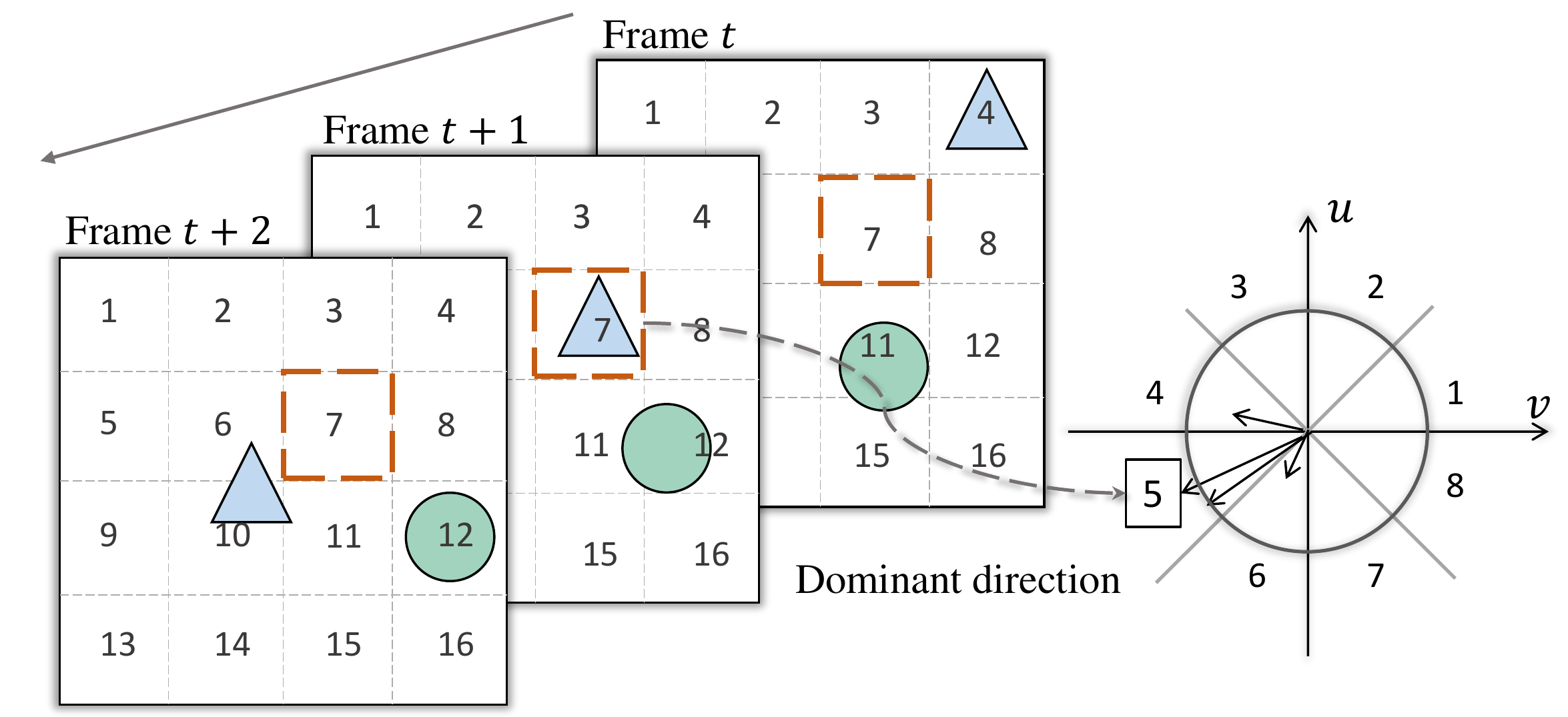}
	\end{center}
	\caption{Illustration of extracting statistical labels in a three-frame video clip. Detailed explanation is in Section~\ref{sec.concept}.}
	\label{fig:concepts}
\end{figure}

Fig. \ref{fig:concepts} shows an example of a three-frame video clip with two moving objects (blue triangle and green circle). 
A typical video clip usually contains much more frames while here we use the three-frame clip example for better understanding. 
To roughly represent the location and quantify ``where'', each frame is divided into $4 \times 4$ blocks and each block is assigned to a number in an ascending order starting from 1 to 16. The blue triangle moves from block 4 to block 7, and the green circle moves from block 11 to block 12. 
By comparing the moving distances, we can easily find that the motion of the blue triangle is larger than the motion of the green circle. 
The largest motion lies in block 7 since it contains moving-in motion between frames~$t$ and~$t+1$, and moving-out motion between frames $t+1$ and $t+2$. 
Regarding the question ``\emph{what is the dominant direction of the largest motion?}'', it can be easily observed that in block 7, the blue triangle moves towards lower-left.
To quantify the directions, the full angle of 360\degree~is divided into eight pieces, with each piece covering a 45\degree~motion direction range, as shown on the right side in Fig.~\ref{fig:concepts}. 
Similar to location quantification, each angle piece is assigned to a number in an ascending order counterclockwise. 
The corresponding angle piece number of ``lower-left'' is 5.

The above illustration explains the basic idea of extracting statistical labels for motion characteristics.
To further consider appearance characteristics ``\emph{where is the largest color diversity and its dominant color?}'', both block 7 and block 12 change from the background color to the moving object color. When considering that the area of the green circle is larger than the area of the blue triangle, we can tell that the largest color diversity location lies in block 12 and the dominant color is green.

Keeping the above ideas in mind, we next formally describe the approach to extract spatio-temporal statistical labels for the proposed pretext task. 
We assume that by training a spatio-temporal CNN to predict the motion and appearance statistics mentioned above, better spatio-temporal representation can be learned, which will benefit the downstream video analysis tasks consequently.

\subsection{Motion Statistics}\label{sec.motionstat}

Optical flow is a commonly used feature to represent motion information in many action recognition methods~\cite{simonyan2014two, carreira2017quo}. 
In the self-supervised learning paradigm, predicting optical flow between every two consecutive frames is leveraged as a pretext task to pre-train the deep model~\cite{gan2018geometry}. 
Here we also leverage optical flow estimated from a conventional non-parametric coarse-to-fine algorithm~\cite{brox2004high} to derive the motion statistical labels that are predicted in our approach.

However, we argue that there are two main drawbacks when directly using dense optical flow to compute the largest motion in our pretext task: (1) optical flow based methods are prone to being affected by camera motion, since they represent the absolute motion \cite{dalal2006human, wang2011action}. (2) Dense optical flow contains sophisticated and redundant information for statistical labels computation, increasing the learning difficulty and leading to network capacity waste. To mitigate these influences, we seek to use a more robust and sparse feature, motion boundary \cite{dalal2006human}.   

\textbf{Motion Boundary.} Denote $u$ and $v$ as the horizontal and vertical components of optical flow, respectively. 
Motion boundaries are derived by computing the x- and y-derivatives of $u$ and $v$, respectively: 
\begin{equation}
\resizebox{0.91\hsize}{!}{
	$	m_u=(u_x, u_y)=(\frac{\partial u}{\partial x}, \frac{\partial u}{\partial y}), ~
	m_v=(v_x, v_y)=(\frac{\partial v}{\partial x}, \frac{\partial v}{\partial y}), $    
}
\label{eq:mb}
\end{equation}
where $m_u$ and $m_v$ is the motion boundary of $u$ and $v$, respectively.
As motion boundaries capture changes in the flow field, constant or smoothly varying motion, such as motion caused by camera view change, will be cancelled out.
Specifically, given an $N$-frame video clip,  $(N-1)\times2$ motion boundaries are computed based on $(N-1)$ optical flows.
Diverse video motion information can be encoded into two summarized motion boundaries by summing up all these $(N-1)$ sparse motion boundaries $m_u$ and $m_v$:
\begin{equation}
M_u=( \sum\limits_{i=1}^{N-1}u_x^i, \sum\limits_{i=1}^{N-1}u_y^i), ~
M_v=( \sum\limits_{i=1}^{N-1}v_x^i, \sum\limits_{i=1}^{N-1}v_y^i),
\label{eq:sum_up}
\end{equation}
where $M_u$ and $M_v$ denotes the summarized motion boundaries on $u$ and $v$, respectively. 

\begin{figure}
	\centering
	\includegraphics[width=\columnwidth]{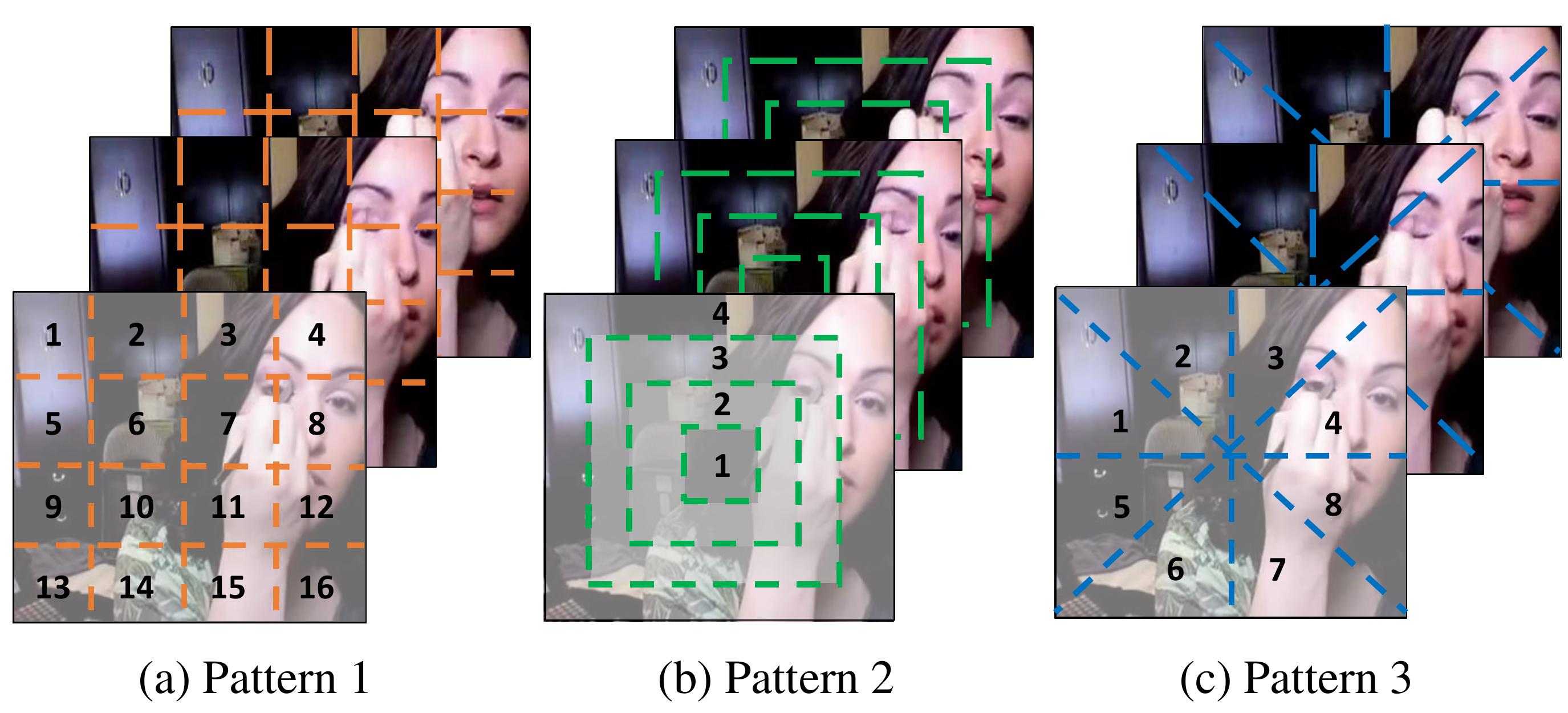}
	\caption{Three different partitioning patterns used to divide video frames into different spatial regions. Each spatial block is assigned with a number to represent its location.}
	\label{fig:pattern}
\end{figure}

\textbf{Spatial-aware Motion Statistical Labels.} Based on motion boundaries, we describe how to compute the spatial-aware motion statistical labels that describe the largest motion location and the dominant direction of the largest motion.
Given a video clip, we first divide it into spatial blocks using partitioning patterns as shown in Fig.~\ref{fig:pattern}. 
Here, we introduce three simple yet effective patterns:
pattern 1 divides each frame into 4$\times$4 grids; pattern 2 divides each frame into 4 different non-overlapped areas with the same gap between each block; pattern 3 divides each frame by two center lines and two diagonal lines. 
Then we compute summarized motion boundaries $M_u$ and $M_v$ as described in Eq.~(\ref{eq:sum_up}).
Motion magnitude and orientation of each pixel can be obtained by transforming $M_u$ and $M_v$ from the Cartesian coordinates to the polar coordinates.  

We take pattern 1 for an example to illustrate how to generate the motion statistical labels, while other patterns follow the same procedure. For the \emph{largest motion location labels}, we first compute the average magnitude of blocks 1 to 16 in pattern 1. Then we compare and find out block~$B$ with the largest average magnitude from the above 16 blocks. The index number of $B$ is treated as the largest motion location label. Note that the largest motion locations computed from $M_u$ and $M_v$ can be different. Therefore, two corresponding labels are extracted from $M_u$ and $M_v$, respectively.

Based on the largest motion block, we compute the \emph{dominant orientation label}, which is similar to the computation of motion boundary histogram~\cite{dalal2006human}. We divide 360\degree~into 8 bins evenly, and assign each bin with a number to represent its orientation.
For each pixel in the largest motion block, we use its orientation angle to determine which angle bin it belongs to and add the corresponding magnitude value into the angle bin. 
The dominant orientation label is the index number of the angle bin with the largest magnitude sum.
Similarly, two orientation labels are extracted from $M_u$ and $M_v$, respectively. 

\wang{\textbf{Global Motion Statistical Labels.} We further propose global motion statistical labels that provide complementary information to the local motion statistics described above. Specifically, given a video clip, the model is asked to predict the frame index (instead of the block index) with the largest motion. To succeed in such a pretext task, the model is encouraged to understand the video contents from a global perspective.}

\wang{Formally, given an $N$-frame video clip, motion boundaries of the $i^{th}$ frame can be computed from Eq.~(\ref{eq:mb}), resulting in $m_u^i$ and $m_v^i$, respectively. 
By casting $m_u^i$ and $m_v^i$ from the Cartesian coordinates to the Polar coordinates, motion magnitude and orientation of each pixel can be obtained.
Denote the magnitude maps as $\mathsf{mag}_u^i$ and $\mathsf{mag}_v^i$. Then the global motion statistical labels can be computed as follows:}
\wang{
\begin{equation}
\resizebox{0.91\hsize}{!}{
$I_u = \underset{i\in \{1,\cdots, N-1\}}{\arg\max} \sum \mathsf{mag}_u^i,~
I_v = \underset{i\in \{1,\cdots, N-1\}}{\arg\max} \sum  \mathsf{mag}_v^i$,
}
\end{equation}
where $I_u$, $I_v$ are the frame indices of the largest magnitude sum (largest motion) w.r.t. $u$ and $v$.}


\subsection{Appearance Statistics}\label{sec.rgbstat}
\wang{\textbf{Spatio-temporal Color Diversity Labels.} Given an  $N$-frame video clip, we divide it into spatial video blocks by patterns described above, same as the motion statistics.   
For an $N$-frame video block, we compute the volumetric color distribution $V_i$ in the 3D color space for the $i^{th}$ frame. 
We then use the Intersection over Union (IoU) along the temporal axis to quantify the spatio-temporal color diversity as follows:}
\wang{
\begin{equation}
\text{IoU}=\frac{V_1\cap V_2 \cap \cdots \cap V_i \cdots \cap V_N}{V_1\cup V_2 \cup \cdots \cup V_i \cdots \cup V_N}.
\end{equation}}
$\!\!\!$The largest color diversity location is the block with the smallest $\text{IoU}$, while the smallest color diversity location is the block with the largest $\text{IoU}$. In practice, we calculate the $\text{IoU}$ on R, G, B channels separately and compute the final $\text{IoU}$ by averaging them as follows:
\wang{
\begin{equation}
\text{IoU}= (\text{IoU}^R+\text{IoU}^G+\text{IoU}^B)~/~3,
\end{equation}
where the 3D distribution $V_i$ used to compute $\text{IoU}^R$, $\text{IoU}^G$, and $\text{IoU}^B$ is generated by calculating the color histogram w.r.t. each color channel.}


\textbf{Dominant Color Labels.} Based on the video blocks with the largest color diversity and smallest color diversity, we compute the corresponding dominant color labels. We divide the 3D RGB color space into 8 bins evenly and assign each bin with an index number. Then for each pixel in the video block, based on its RGB value, we assign a  number of the corresponding color bin to it. Finally, color bin with the largest number of pixels is the label for the dominant color. 

\textbf{Global Appearance Statistical Labels.} We also propose global appearance statistical labels to provide supplementary information. Specifically, we use the dominant color of the whole video (instead of a video block) as the global appearance statistical label. The computation method is the same as the one described above.  

\begin{figure*}
	\centering
	\includegraphics[width=0.95\linewidth]{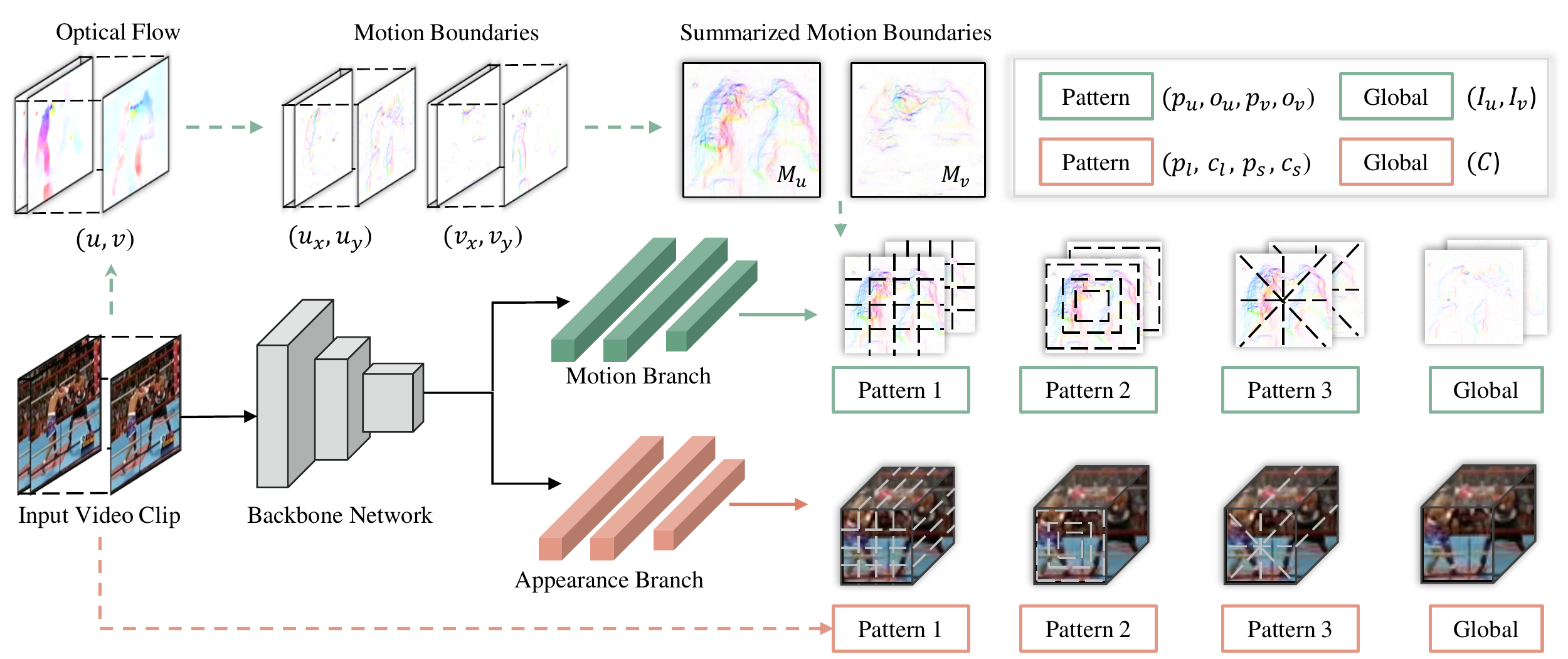}
	\caption{\wang{Framework of the proposed approach. Given a video clip, 14 motion statistical labels and 13 appearance statistical labels are to be predicted.} The motion statistical labels are computed from the summarized motion boundaries $M_u$ and $M_v$. The appearance statistical labels are computed from the input video clip. For each local motion pattern, 4 ground-truth labels are generated: $p_u$, $o_u$---the spatial location of the largest magnitude based on $M_u$ and its corresponding dominant orientation; $p_v$, $o_v$---the spatial location of the largest magnitude based on $M_v$ and its corresponding dominant orientation. Two global motion statistical labels are $I_u$, $I_v$---the frame indices of the largest magnitude sum w.r.t. $m_u$ and $m_v$. For each local appearance pattern, 4 ground-truth labels are generated: $p_l$, $c_l$---the spatial location of the largest color diversity and its corresponding dominant color; $p_s$, $c_s$---the spatial location of the smallest color diversity and its corresponding dominant color. The global appearance statistical label is $C$---the dominant color of the whole video.}
	\label{fig:network}
\end{figure*}

\subsection{\wang{Motion-Aware Curriculum Learning}} \label{sec:cl}

We further propose to leverage the curriculum learning strategy~\cite{bengio2009curriculum} to improve the learning performance. The key concept is to present the network with more difficult samples gradually. It is inspired by the human learning process and proven to be effective on many learning tasks \cite{han2019video, korbar2018cooperative, sumer2017self}. Recently, Hacohen and Weinshall \cite{hacohen2019power} further investigated the curriculum learning in training deep neural networks and proposed two fundamental problems to be solve: (1) scoring function problem, \ie, {how to quantify the difficulty of each training sample}; (2) pacing function problem, \ie, {how to feed the networks with the sorted training samples.} 
In this work, for self-supervised video representation learning, we describe our solutions to these two problems as follows.

\textbf{Scoring Function.} Scoring function $f$ defines how to measure the difficulty of each training sample. In our case, each video clip is considered to be easy or hard, based on the difficulty to figure out the block with the largest motion, \ie, difficulty to predict the motion statistical labels.
To characterize the difficulty, we use the ratio between magnitude sum of the largest motion block and magnitude sum of the entire videos, as the scoring function $f$. When the ratio is large, it indicates that the largest motion block contains the dominant action in the video and thus is easy to find out the largest motion location, \eg, a man skiing in the center of a video with smooth background change. On the other hand, when the ratio is small, it indicates that the action in the video is relatively diverse or the action is less noticeable, \eg, two persons boxing with another judge walking around. See Section~\ref{sec:CL_exp} for more visualized examples.

Formally, given an $N$-frame video clip, two summarized motion boundaries $M_u$ and $M_v$ are computed based on Eq.~(\ref{eq:sum_up}) and the corresponding magnitude maps are denoted by $M_u^{mag}$ and $M_v^{mag}$. Denote  $B_u$, $B_v$ as the largest motion blocks, and $B_u^{mag}$, $B_v^{mag}$ as the corresponding magnitude maps. The scoring function $f$ is defined as the maximum ratio between the magnitude sum of $B_u$, $M_u$ and $B_v$, $M_v$:

\begin{equation}
f=\max(~\frac{\sum B_u^{mag}}{\sum M_u^{mag}},~
\frac{\sum B_v^{mag}}{\sum M_v^{mag}}).
\end{equation}
Here we use the maximum ratio between the horizontal component $u$ and the vertical component $v$ as the difficulty score. The reason is that large magnitude in \emph{one} direction can already define large motion, \eg, a person running from left to right contains large motion in horizontal direction~$u$ but small motion in vertical direction $v$. With the scores computed from function $f$, training samples are sorted in a descending order accordingly, representing the difficulty from easy to hard.

\textbf{Pacing Function.} After sorting the samples, the remaining question is how to split these samples into different training steps. Previous works~\cite{han2019video, korbar2018cooperative, sumer2017self} usually adopt a two-stage training scheme, \ie, training samples are divided into two categories: easy and hard. In \cite{hacohen2019power}, the authors formally define such a problem as a pacing function~$g$, and introduce three stair-case functions: \emph{single step}, \emph{fixed exponential pacing}, and \emph{varied exponential pacing}, where they demonstrate that these functions have comparable performances~\cite{hacohen2019power}. In our case, we adopt the simple single step pacing function (we also tried other functions and similarly found that they show comparable performances). 
\wang{Formally, we denote the first half of the training \jb{samples} (descendingly sorted as aforementioned) as $S_1$ and the remaining half \jb{samples} as~$S_2$. Then the pacing function is defined as follows:}
\wang{
\begin{equation}
    g(i) = S_1 + H(i-t) \cdot S_2,
\end{equation}
where $i$ is the training iteration, $H(i-t)$ is the Heaviside step function~\cite{davies2002integral}, and $t$ is 
the number of iterations when the remaining half \jb{samples} $S_2$ are included for pre-training.
In practice, the entire $S$ will then be used for the second-stage training when the model is converged on the first half training samples. }

\makeatletter
\renewcommand\@seccntformat[1]{\color{black} {\csname the#1\endcsname}\hspace{0.5em}}
\makeatother

\subsection{Learning with Spatio-temporal CNNs}\label{sec.learnc3d}

The framework of our approach is illustrated in Fig. \ref{fig:network}. 
Given an input video clip, a neural network is trained to \jb{uncover} motion and appearance statistic\jb{s defined above}.   
Specifically, we consider C3D \cite{tran2015learning}, 3D-ResNet \cite{hara2018can}, R(2+1)D\cite{tran2018closer}, and S3D-G~\cite{xie2018rethinking} as \emph{Backbone Networks}. Regarding training targets, \jb{the proposed task could be modeled as} either a regression problem or a classification problem. 
The effectiveness of different backbone networks \jb{and} training targets is thoroughly \jb{analyzed} in Sections~\ref{sec:bakcbone_exp} and \ref{sec:target_exp}, respectively.

\subsubsection{\wang{Backbone Network}}
\emph{C3D}~\cite{tran2015learning} network extends 2D convolutional kernel $k \times k$ to 3D convolutional kernel $k \times k \times k$ to operate on  3D video volumes. It contains 5 convolutional blocks, 5 max-poling layers, 2 fully-connected layers, and a soft-max layer in the end to predict action class. Each convolutional block contains 2 convolutional layers except the first two blocks. Batch normalization (BN) is also added between each convolutional layer and ReLU layer.   

\emph{3D-ResNet}~\cite{hara2018can} is a 3D extension of the widely used 2D architecture ResNet~\cite{he2016deep}, which introduces shortcut connections that perform identity mapping of each building block. A basic residual block in 3D-ResNet (R3D) contains two 3D convolutional layers with BN and ReLU followed. Shortcut connection is introduced between the top of the block and the last BN layer in the block. Following previous work \cite{hara2018can}, we use 3D-ResNet18 (R3D-18) as our backbone network, which contains four basic residual blocks and one traditional convolutional block on the top.

\emph{R(2+1)D}~\cite{tran2018closer} 
breaks the original spatio-temporal 3D convolution into a 2D spatial convolution and a 1D temporal convolution, since 3D CNNs are computationally expensive.
While preserving similar network parameters to R3D, R(2+1)D outperforms R3D on supervised video action recognition task. 

\wang{\emph{S3D-G}~\cite{xie2018rethinking} builds on top of I3D~\cite{carreira2017quo}. While achieving  comparable results with I3D, S3D-G is much more efficient by introducing three vital ideas. First, it adopts a top-heavy design, \ie, uses 2D convolutions in the lower layers and 3D convolutions in the higher layers. Second, it replaces 3D convolutions with separable 2D and 1D convolutions. Third, it uses spatio-temporal feature gating.} 

\subsubsection{\wang{Training Targets}}

\wang{\textbf{Regression.}
We \jb{first} model the spatio-temporal statistics prediction task as a regression problem. Specifically, we consider two different designs: 1D label regression and 2D label regression.}

\wang{In the 1D label regression design, we represent each statistical label by a 1D label. In this case, 27 values, including 14 motion statistical labels and 13 appearance statistical labels, are to be regressed. 
In the 2D label regression design, we represent the spatial location of patterns 1 and 3 by a 2D label. 
For example, in pattern 1, block 1 will be represented as $(1,1)$ and block 7 will be represented as $(2,3)$.
In this case, 35 values, including 18 motion statistical labels and 17 appearance statistical labels, are to be regressed.}
\wang{$L_2$-norm is leveraged as the loss function to measure the difference between predicted labels and target labels. Formally, the loss function is defined as:
\begin{equation}
\mathcal{L}_{reg} = \lambda_{m}\lVert \hat{y}_{m} - y_{m} \rVert_2 + \lambda_{a}\lVert \hat{y}_{a} - y_{a} \rVert_2,
\label{eq:reg_loss}
\end{equation}
where $\hat{y}_{m}$, $y_m$ denote the predicted and target motion statistical labels, and $\hat{y}_{a}$, $y_a$ denote the predicted and target appearance statistical labels.  
$\lambda_m$ and $\lambda_a$ are the weighting parameters that are used to balance the two loss terms.}

\wang{\textbf{Classification.}
We further consider modeling our pretext task as a classification problem.
Each statistical label will be predicted independently by a fully-connected layer with a cross-entropy loss. 
In this case, it will introduce 27 fully-connected layers.
\jb{Although optimizing through such a large number of fully-connected layer seems to bias the learning towards these layers instead of the preceding layers, our experimental analysis (Section~\ref{sec:target_exp}) reveals some different findings.}
Formally, the loss function is defined as:
\begin{equation}
\mathcal{L}_{cls} = \lambda_{m}\sum\limits_{i=1}^{\jb{n_m}}\mathcal{L}_{m\_cls}^i +\lambda_{a}\sum\limits_{j=1}^{\jb{n_a}}\mathcal{L}_{a\_cls}^j,
\end{equation}
where \jb{$n_m$ and $n_a$} denote the numbers of motion and appearance statistical labels \jb{($n_m=14$ and $n_a=13$ in our paper)}. Specifically,
\begin{equation}
\begin{aligned}
\mathcal{L}_{m\_cls}^i = -\sum\limits_{c=1}^{M_i} y_{o,c}^m \log (p_{o,c}^m),\\ \mathcal{L}_{a\_cls}^j = -\sum\limits_{c=1}^{M_j} y_{o,c}^a \log (p_{o,c}^a),
\end{aligned}
\end{equation}
where $\mathcal{L}_{m\_cls}^i$, $\mathcal{L}_{a\_cls}^j$ denote the $i^{th}$ motion and $j^{th}$ appearance classification losses,
$M_i$, $M_j$ denote the number of classes for the $i^{th}$ motion and $j^{th}$ appearance statistical labels,
$y_{o,c}^m$, $y_{o,c}^a$ indicate whether the predicted label $c$ is the same as the target label $o$, 
and $p_{o,c}^m$, $p_{o,c}^a$ denote the predicted probabilities.
$\lambda_m$ and $\lambda_a$ are the weighting parameters that are used to balance the two loss terms.}

\makeatletter
\renewcommand\@seccntformat[1]{\color{black} {\csname the#1\endcsname}\hspace{0.5em}}
\makeatother

\section{Experimental Setup}\label{sec:setup}

\subsection{Datasets}

We conduct extensive experimental evaluations on five datasets in the following sections. 

\emph{Kinetics-400}~(K-400)~\cite{kay2017kinetics} 
contains around 306k videos of 400 action classes. It is divided into three splits: training split, validation split, and testing split. Following previous works~\cite{han2019video, wang2020self, benaim2020speednet}, we use the training split as pre-training dataset, which contains around 240k video samples.

\emph{UCF101}~\cite{soomro2012ucf101} is a widely used dataset which contains 13,320 video samples of 101 action classes. It is divided into three splits. Following previous works~\cite{han2019video, xu2019self, wang2020self}, we use the \emph{training split 1} as pre-training dataset and the \emph{training/testing split 1} for downstream task evaluation.

\emph{HMDB51}~\cite{kuehne2011hmdb} is a relatively small action dataset which contains around 7,000 videos of 51 action classes. 
Following previous works~\cite{han2019video, xu2019self, wang2020self}, we use the \emph{training/testing split 1} to evaluate the proposed approach.  

\emph{YUPENN} \cite{derpanis2012dynamic} is a dynamic scene recognition dataset which contains 420 video samples of 14 dynamic scenes. We follow the recommended leave-one-out evaluation protocol~\cite{derpanis2012dynamic} when evaluating the proposed approach. 

\emph{ASLAN} \cite{kliper2011action} is a video dataset focusing on the action similarity labeling problem and contains 3,631 video samples of 432 classes. 
During testing, following previous work~\cite{kliper2011action}, we use a 10-fold cross validation with leave-one-out evaluation protocol. 

\subsection{Implementation Details}

\textbf{Self-supervised Pre-training Stage.} When pre-training on UCF101 dataset, video samples are first split into non-overlapped 16 frame video clips and are randomly selected during pre-training. When pre-training on K-400, following previous works~\cite{kim2018self, han2019video}, we randomly select a consecutive 16-frame video clip and the corresponding 15-frame optical flow clip from each video sample. 
Each video clip is reshaped to spatial size of 128 $\times$ 171. As for data augmentation, we randomly crop the video clip to 112 $\times$ 112 and apply random horizontal flip for the entire video clip. Weights of motion statistics $\lambda_m$ and appearance statistics $\lambda_a$ are empirically set to be 1 and 0.1. The batch size is set to 30. We use Stochastic Gradient Descent (SGD) as the optimizer with learning rate $5\times10^{-4}$, which is divided by 10 every 6 epochs and the training process is stopped at 25 epochs. Model with the lowest validation loss is used for downstream stream video analysis tasks.  

\textbf{Supervised Fine-tuning Stage.} During the supervised fine-tuning stage, weights of convolutional layers are retained from the self-supervised pre-trained models and weights of the fully-connected layers are re-initialized. The whole network is then trained again with cross-entropy loss on action recognition task with UCF101 and HMDB51 datasets. Image pre-processing procedure and training strategy are the same as the self-supervised pre-training stage, except that the initial learning rate is changed to $3\times10^{-3}$. 

\textbf{Evaluation.} For action recognition task, during testing, video clips are resized to 128 $\times$ 171 and center-cropped to 112 $\times$ 112. We consider two evaluation methods: clip accuracy and video accuracy. The clip accuracy is computed by averaging the accuracy of each clip from the testing set.  The video accuracy is computed by averaging the softmax probabilities of uniformly selected clips in each video~\cite{xu2019self} from the testing set. In all of the following experiments, to have a fair comparison with previous works~\cite{xu2019self, luo2020video, han2019video}, we use video accuracy to evaluate our approach, apart from the ablation studies on the effectiveness of each component (Section \ref{sec:component_exp}), where we use clip accuracy to keep a consistency with our previous conference paper~\cite{wang2019self}. 

We further use our self-supervised pre-trained models as feature extractors 
on three downstream video analysis tasks: dynamic scene recognition, and action similarity labeling. More evaluation details are presented in Section~\ref{sec:sota} for every downstream task.

\section{Ablation Studies and Analyses}\label{sec:ablation}

\wang{In this section, we conduct extensive ablation studies to analyze the proposed approach. 
We study the effectiveness of each component in  Section~\ref{sec:component_exp}, the effectiveness of different backbone networks in Section~\ref{sec:bakcbone_exp}, the correlation between 
pretext and downstream task performances
in  Section~\ref{sec:pre_vs_down_exp}, the effectiveness of the pre-training dataset in Section~\ref{sec:data_exp}, and the effectiveness of curriculum learning strategy in Section~\ref{sec:CL_exp}.
All these studies are conducted on top of the 1D label regression design.
Then we investigate the effectiveness of different training targets in Section~\ref{sec:target_exp}.} 
\begin{table*}[htbp!]
    \small
    \renewcommand{\tabcolsep}{8.0pt}
	\captionof{table}{Ablation Experiments on Spatio-temporal Statistics Component.}
	\begin{tabular}{ccc}
	(a) Partitioning statistical patterns &
	(b) Local and global statistics &
	(c) Motion and appearance statistics \\
	  \begin{tabular}{lcc}
			\toprule
			\renewcommand{\tabcolsep}{6.0pt}
			Initialization & UCF101 \\
			\midrule
			Random & 45.4 \\
			Motion pattern 1 & 53.8 \\
			Motion pattern 2 & 53.2 \\
			Moiton pattern 3 & 54.2\\
			\bottomrule
		\end{tabular}
		& \begin{tabular}{lc}
			\toprule
			Initialization & UCF101 \\
			\midrule
			Random & 45.4 \\
			Motion global & 48.3 \\
			Motion pattern all & 55.4\\
			Motion pattern all + global & 57.8\\
			\bottomrule
		\end{tabular} 
	     & 
	     \begin{tabular}{lcc}
				\toprule
				Initialization & UCF101 & HMDB51\\
				\midrule
				Random & 45.4 & 19.7\\
				Appearance & 48.6 & 20.3\\
				Motion & 57.8 & 30.0\\
				Joint & 58.8 & 32.6 \\
				\bottomrule
			\end{tabular}
	\end{tabular}
\label{tab:cpn}
\end{table*}


\begin{table*}[tbp!]
	\begin{minipage}{.49\textwidth}
		\caption{
		Evaluation of different backbone networks on UCF101 and HMDB51 datasets.
		When pre-training, we use our self-supervised pre-training model as weight initialization. 
		}
	    \begin{center}
	   \small
	   \begin{adjustbox}{max width =\linewidth}
		\begin{tabular}{ccccc}
			\toprule
			\multicolumn{3}{c}{Experimental setup} &
			\multicolumn{2}{c}{Downstream task} \\
			\cmidrule(lr){1-3}
			\cmidrule(lr){4-5}
			Pre-training & Backbone &  \#Params. &  UCF101 & HMDB51  \\
			\midrule
			\xmark & C3D  & 33.4M & \textbf{61.7} & \textbf{24.0} \\
			\checkmark & C3D  & 33.4M  &  69.3 & 34.2 \\
			\midrule
			\xmark & R3D-18  & 14.4M  &  54.5 & 21.3\\
			\checkmark & R3D-18& 14.4M &  67.2 & 32.7 \\
			\midrule
			\xmark & R(2+1)D  & 14.4M &  56.0 & 22.0\\
			\checkmark & R(2+1)D  & 14.4M &  \textbf{73.6} & \textbf{34.1} \\
			\bottomrule
		\end{tabular}
	\end{adjustbox}
	\end{center}
	\label{tab:backbone}
	\end{minipage}%
	\hfill
	\begin{minipage}{.49\textwidth}
		\centering\includegraphics[width=1\linewidth]{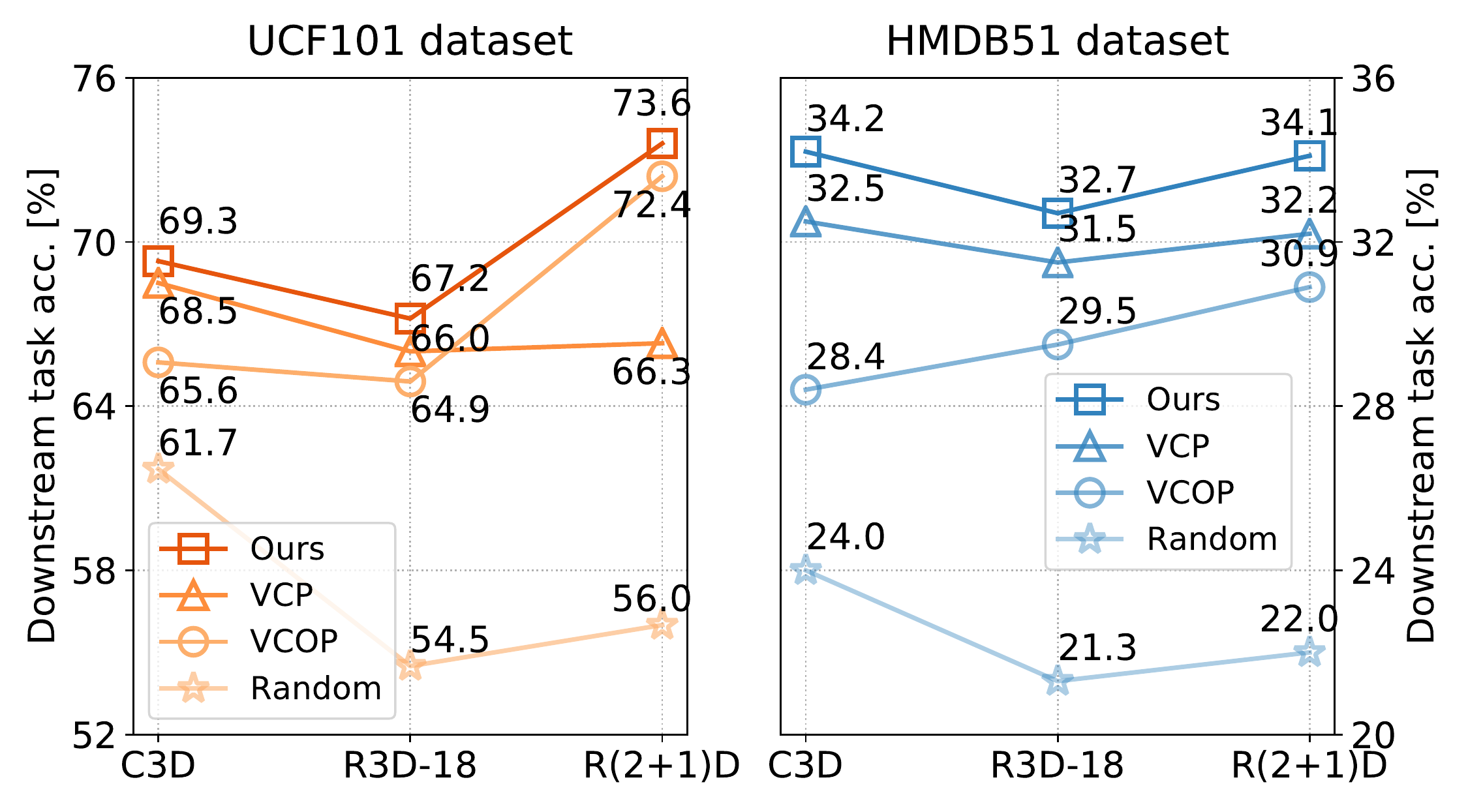}
		\captionof{figure}{
		Action recognition accuracy in terms of four initialization methods w.r.t three backbone networks on UCF101 and HMDB51 datasets.
		}
		\label{fig:backbone}
	\end{minipage}
\end{table*}

\subsection{Effectiveness of Each Component} \label{sec:component_exp}

\textbf{Pattern.} We study the performances of three partitioning patterns.
Here, we analyze and show the performances based on the motion statistics while the appearance statistics follows
the same trend. As shown in Table~\ref{tab:cpn}(a), all three patterns achieve comparable
results. 
Compared to random initialization, \ie, training from scratch, each pattern improves by around 8\%. 

\textbf{Local vs. Global.} We study the performances of local statistics, \textit{where is the largest motion location?},  global statistics, \textit{which is the largest motion frame?}, and their ensemble. As can be seen in Table \ref{tab:cpn}(b), when the three local patterns are combined together, we can further get around 1.5\% improvement, compared to single pattern in Table~\ref{tab:cpn}(a). The global statistics also serves as a useful supervision signal with an improvement of 3\%. All motion statistical labels, \ie, local and global statistics, 
achieve 57.8\% accuracy on the UCF101 dataset.

\textbf{Motion, RGB, and Joint Statistics.} We finally analyze the performances of motion statistics, appearance statistics, and their combination in Table \ref{tab:cpn}(c). Both appearance and motion statistics serve as useful self-supervised signals but the motion statistics is more powerful. We hypothesize the reason is that temporal information could be more important for action recognition task. When combining motion and appearance statistics, the action recognition accuracy can be further improved.


\subsection{Effectiveness of Backbone Networks}\label{sec:bakcbone_exp}

Recently, modern spatio-temporal representation learning architectures, such as R3D-18~\cite{hara2018can} and R(2+1)D~\cite{tran2018closer}, have been used to validate self-supervised video representation learning methods~\cite{luo2020video, xu2019self}. While the performances of downstream tasks are significantly improved, this practice introduces a new 
variable, 
backbone network, which could interfere with the evaluation of the pretext task itself. In the following, we first evaluate our proposed method with these modern backbone networks in Table \ref{tab:backbone}. Following that, we compare our method with recent works~\cite{luo2020video, xu2019self} on these three 
backbone networks
in Fig.~\ref{fig:backbone}.

We present the performances of different backbone networks on UCF101 and HMDB51 datasets under two settings: without per-training and with pre-training in Table \ref{tab:backbone}. When there is no pre-training, baseline results are obtained by training from scratch. When there is pre-training, backbone networks are first pre-trained on UCF101 dataset with the proposed method and then used as weights initialization for the following fine-tuning. 
We have the following observations: (1) 
Significant
improvement is achieved on both action recognition datasets across three backbone networks. With C3D it improves UCF101 and HMDB51 by 9.6\% and 13.8\%;  with R3D-18 it improves UCF101 and HMDB51 by 13.6\% and 12.1\%; with R(2+1)D it improves UCF101 and HMDB51 by 19.5\% and 15.9\% remarkably. 
(2) Compared to C3D, R3D-18 and R(2+1)D benefit more from the self-supervised pre-training. 
While
C3D achieves the best performance in the no pre-training setting, R(2+1)D finally achieves the highest accuracy on both datasets in the self-supervised setting. (3) The proposed method using 
R(2+1)D
achieves better performance than using 
R3D-18,
while with similar number of network parameters. Similar observation is also demonstrated in supervised action recognition task~\cite{tran2018closer}, where R(2+1)D performs better than R3D-18 on K-400 dataset. 

We further compare our method with two recent proposed pretext tasks VCOP~\cite{xu2019self} and VCP~\cite{luo2020video} on these three backbone networks in Fig.~\ref{fig:backbone}. 
We have three key observations:
(1) The proposed self-supervised learning method achieves the best performance across all three backbone networks on both UCF101 and HMDB51 datasets. This demonstrates the superiority of our method and shows that the performance improvement is not merely due to the usage of the modern networks. 
The proposed spatio-temporal statistical labels indeed drive neural networks to learn powerful spatio-temporal representation for action recognition. (2) For all three pretext tasks, R(2+1)D 
achieves
the largest improvement 
on both datasets, which is similar to the observation in the above experiments. (3) No best network architecture is guaranteed for different pretext tasks. R(2+1)D achieves the best performance with our method and VCOP, while C3D achieves the best performance with VCP.

\subsection{\jb{Pretext Task \emph{vs.} Downstream Task}} \label{sec:pre_vs_down_exp}

\begin{figure}[htbp!]
    \centering
    \includegraphics[width=1\linewidth]{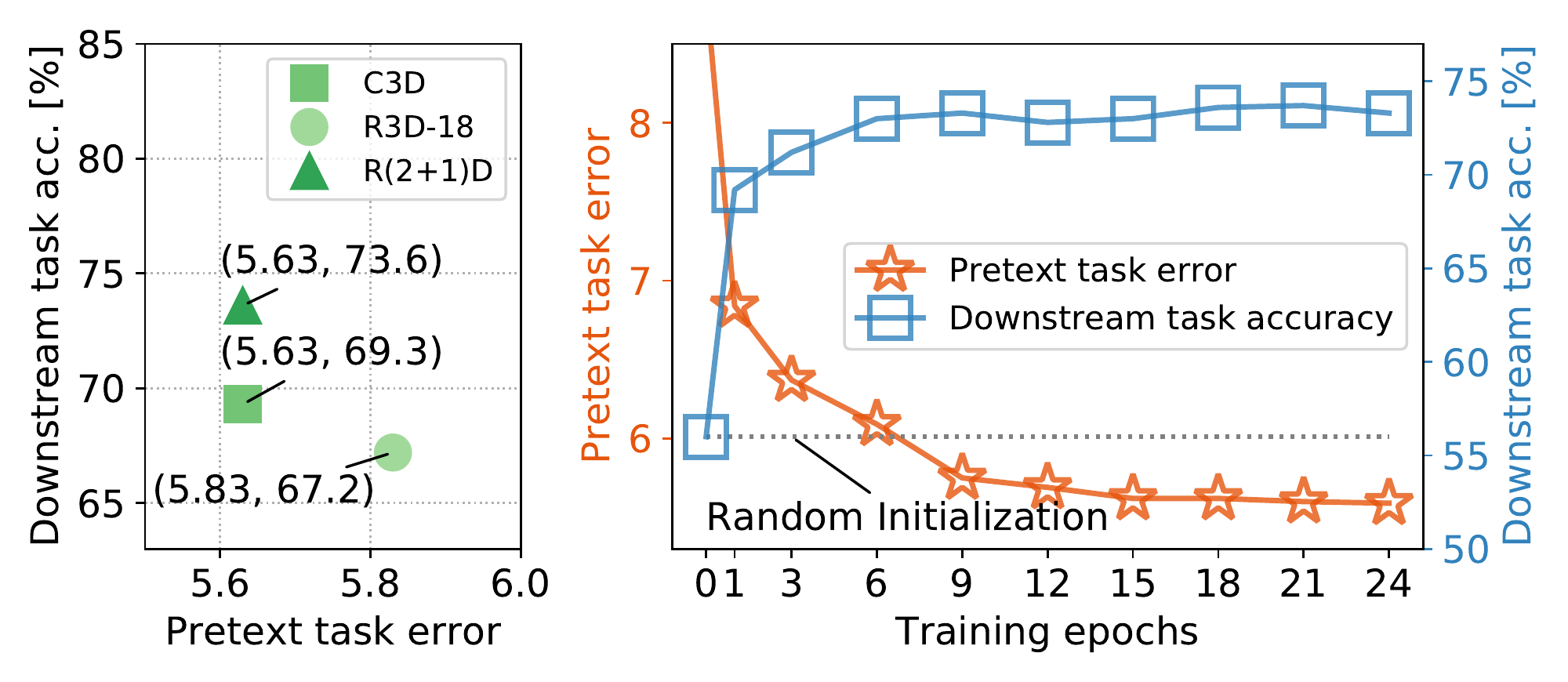}
    \caption{
    Correlation between pretext task error and downstream task accuracy.
    Left: representative value obtained from the best performed model using different backbone networks. Right: Complete evolution using R(2+1)D as the backbone network.}
	\label{fig:pre_vs_down}
\end{figure}

\wang{
We show the correlation between pretext and downstream task performances in Fig.~\ref{fig:pre_vs_down}. Specifically, we use UCF101 training split 1 as the pre-training dataset. The pretext task performance is evaluated by the mean square error between the target and predicted spatio-temporal statistical labels on UCF101 testing split 1. A lower pretext task error indicates a better pretext task performance. The downstream task performance is evaluated by action recognition accuracy on the UCF101 dataset.}

\wang{We have the following observations: 
(1) When using different backbone networks, 
a better pretext task performance does not guarantee a better downstream task performance.
For example, in the left of Fig.~\ref{fig:pre_vs_down}, while C3D and R(2+1)D produce comparable pretext task error, R(2+1)D outperforms C3D by 4.3 \%.       
(2) On the contrary, when the backbone network is fixed to R(2+1)D, as shown in the right of Fig~\ref{fig:pre_vs_down}, with the pretext task error decreasing, action recognition accuracy on UCF101 dataset increases.
(3) The first few pre-training epochs play an important role in the downstream task performance improvement. For example, training 3 epochs can already lead to a significant improvement in the downstream task performance. 
}

\makeatletter
\renewcommand\@seccntformat[1]{\color{black} {\csname the#1\endcsname}\hspace{0.5em}}
\makeatother

\subsection{Effectiveness of Pre-training Data} \label{sec:data_exp}
In the following, we consider two scenarios to investigate the effectiveness of pre-training data. One is the comparison on different pre-training datasets with different data scales. The other is the comparison on the same pre-training dataset but with different sizes of pre-training data.

\textbf{Pre-training Dataset Analysis.} We analyze the performances of training on a relatively small-scale dataset UCF101~\cite{soomro2012ucf101} and 
a large-scale dataset K-400~\cite{kay2017kinetics}. The pre-trained models are evaluated on UCF101 and HMDB51 datasets w.r.t. three different backbone networks. 
As shown in Fig. \ref{fig:acc_vs_dataset}, the performance can be further improved when pre-training on a larger dataset w.r.t all the backbone networks on both downstream datasets.

\begin{table*}[tbp!]
	\begin{minipage}{.5\textwidth}
    \includegraphics[width=1\linewidth]{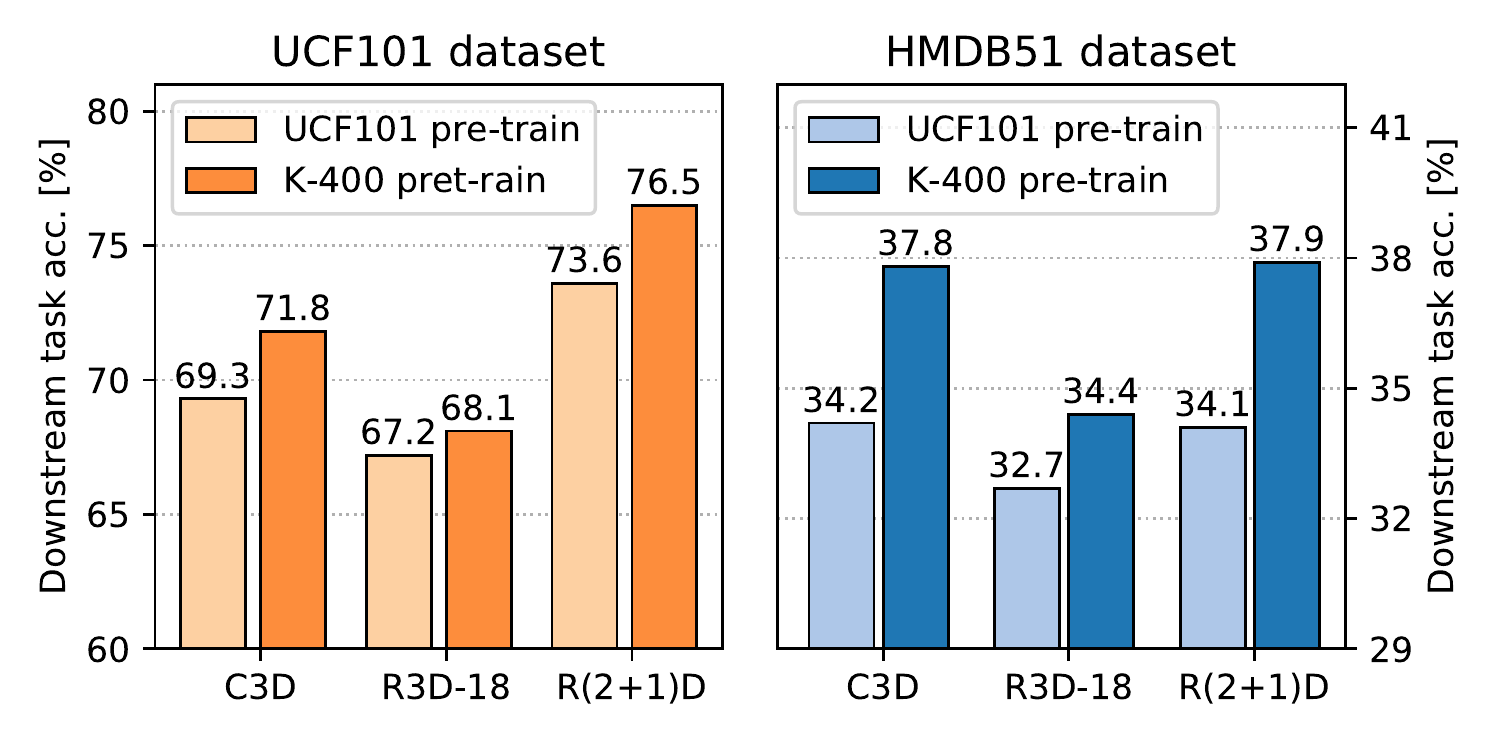}
    \captionof{figure}{ Action recognition accuracy in terms of different pre-training datasets w.r.t. three backbone networks on UCF101 and HMDB51.
    }
    \label{fig:acc_vs_dataset}		
	\end{minipage}%
	\hfill
	\begin{minipage}{.49\textwidth}
    \caption{Evaluation of curriculum learning strategy. ``$\uparrow$'' represents the first half of the K-400 dataset while ``$\downarrow$'' indicates the last half of the K-400 dataset.}
    \centering
	    \begin{adjustbox}{max width=\columnwidth}
	    \renewcommand{\arraystretch}{1.2}
		\begin{tabular}{cccc}
			\toprule
			\multicolumn{2}{c}{Experimental setup} &
			\multicolumn{2}{c}{Downstream task} \\
			\cmidrule(lr){1-2}
			\cmidrule(lr){3-4}
			 Curr. Learn. & Pre-training data  & UCF101 & HMDB51  \\
			\midrule
			\xmark &100 \% K-400 & 76.5 & 37.9\\
			\midrule
			\xmark &50 \% K-400  & 73.6 & 35.6\\
			\xmark &$\uparrow$, 50\% K-400 (simple)  & 72.4 & 35.9\\
			\xmark &$\downarrow$, 50\% K-400 (difficult) & 72.8 & 32.1 \\
			\midrule
			\checkmark &100\% K-400 & \textbf{77.8} & \textbf{40.5}\\
			\bottomrule
		\end{tabular}
	\end{adjustbox}
	\label{tab:cl_results}
	\end{minipage}
\end{table*}

\begin{figure*}[htbp]
\footnotesize
\begin{tabular}{cc}
     \includegraphics[width=1\columnwidth]{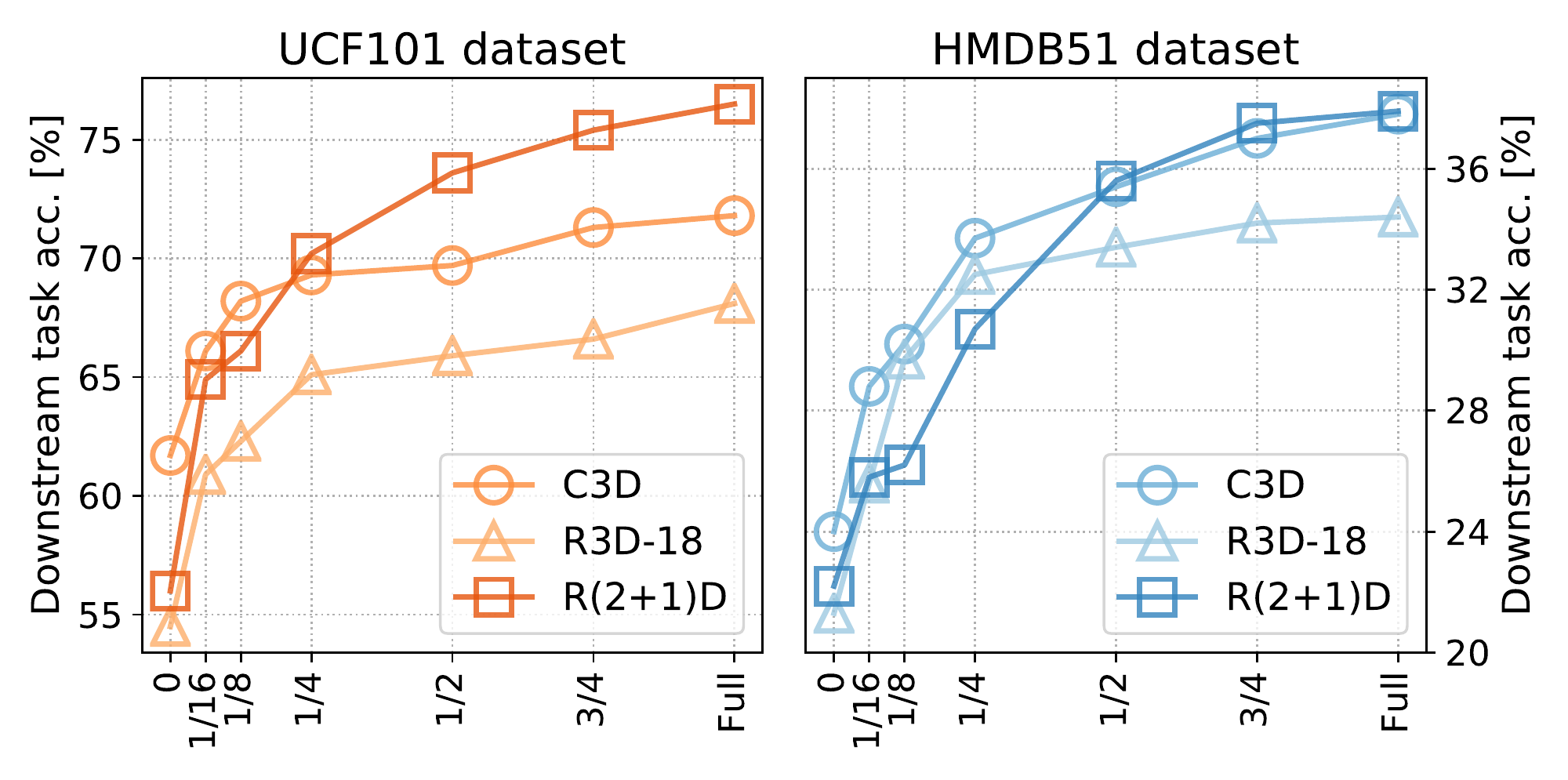} &
     \includegraphics[width=1\columnwidth]{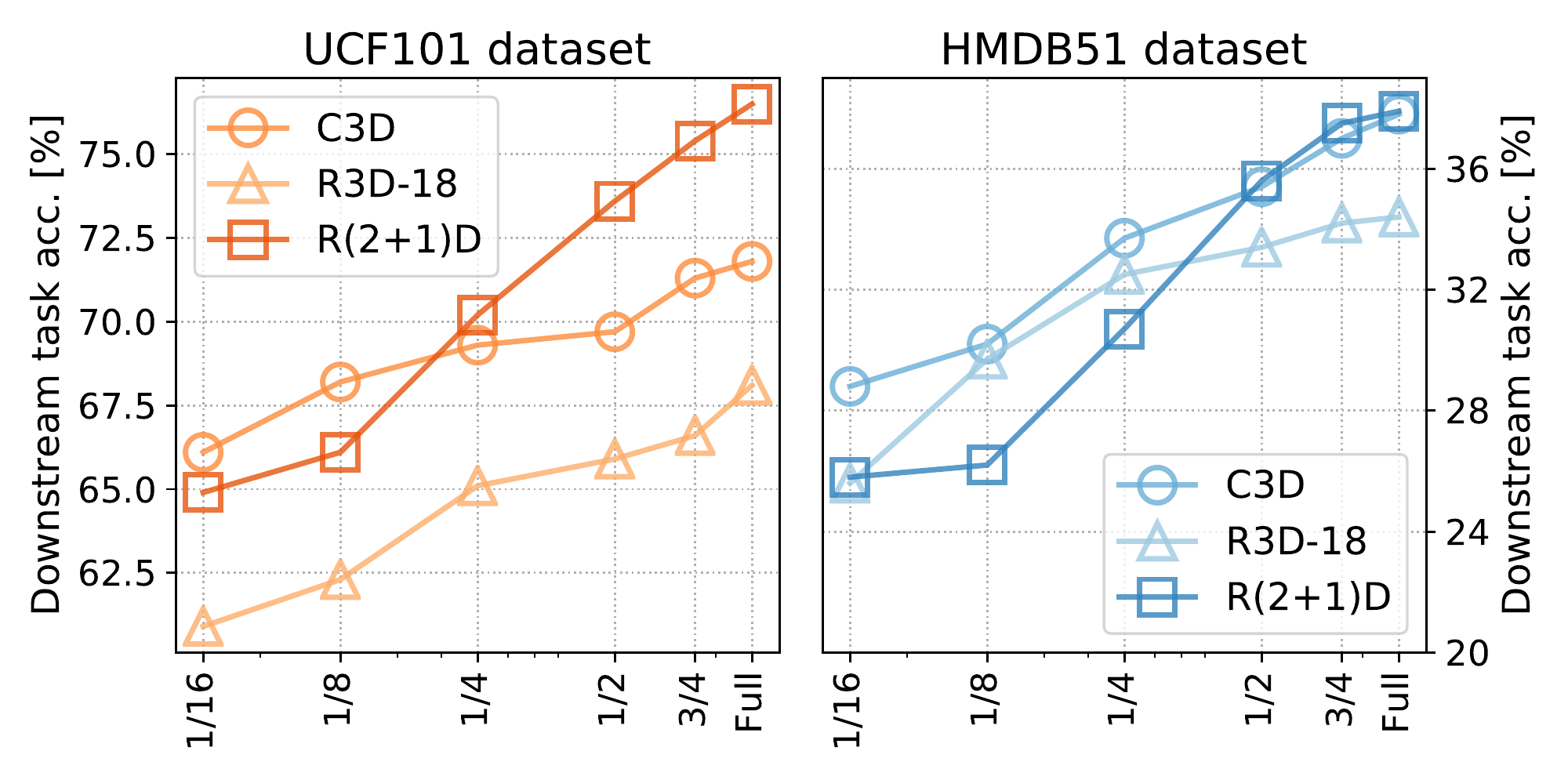} \\
	 (a) & (b)
\end{tabular}
  \caption{Action recognition accuracy in terms of different pre-training dataset scales of K-400 shown in (a) linear-scale (position ``0'' indicates random initialization), and (b) log-scale w.r.t three backbone networks on UCF101 and HMDB51 datasets.}
  \label{fig:data_scale}
\end{figure*}

\textbf{Dataset Scale Analysis.} We further consider pre-training networks on different proportions of \emph{the same} K-400 dataset.  In practice, $1/k$ of K-400 is used for pre-training, where $k= 16, 8, 4, 2, 4/3, 1$.  To obtain the corresponding pre-training dataset, for $k=16, 8, 4, 2$, we select one sample from every $k$ samples of the original full K-400. As for $k=4/3$, we first retain half of the K-400, and then select one sample from every 2 samples in the remaining half dataset.
We conduct extensive experiments on three backbone networks and two downstream datasets.

\wang{As shown in  Fig.~\ref{fig:data_scale}(a), the increase of pre-training data scale does not lead to a linear increase of performance.
Specifically, the increasing speed of performance is high at the beginning and then gradually decreases. 
Let us take R(2+1)D for example. Using $1/8$ of the K-400 can achieve half of the improvement compared to training from scratch.  
In addition, compared with using full K-400, using half of the K-400 only leads to \emph{inconsequential} drop from the highest performance. 
When the $x$-axis is shown in log-scale as in Fig.~\ref{fig:data_scale}(b), with the increase of the pre-training data size, the performance of the proposed method improves \textit{log-linearly}.
Similar observations on this log-linearly improvement property are also reported in other pretext tasks of self-supervised visual representation learning~\cite{goyal2019scaling} and in supervised learning~\cite{ghadiyaram2019large,sun2017revisiting,huh2016makes}.}

\subsection{Effectiveness of Curriculum Learning Strategy} \label{sec:CL_exp}
\wang{
We have shown that downstream task performances improve \textit{log-linearly} with the pre-training data size in Section~\ref{sec:data_exp}. }
\wang{Based on this \jb{observation}, we suggest that it is an interesting direction to investigate the importance of different training samples in improving the downstream task performance. 
In this way, we can utilize the large amount of video data for self-supervised learning in a more efficient way. 
In the following, we show that by using the proposed curriculum learning strategy, the performance can be further improved using the same full K-400 as the pre-training dataset.}

\begin{figure*}[htp!]
	\begin{center}
		\includegraphics[width=\textwidth]{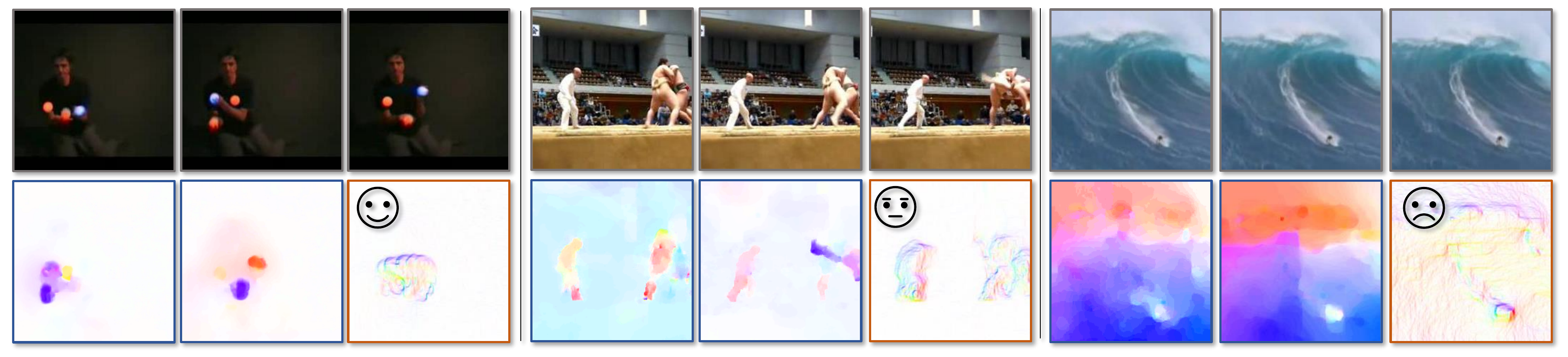} 
	\end{center}
	\caption{Three video samples of the curriculum learning strategy. From left to right, the difficulty to predict the motion statistical labels of each video clip is increasing. For each sample, the top three images are the first, middle, and last frames of a video clip. In the bottom row, the first two images are the corresponding optical flows and the last image is the summarized motion boundaries $M_u/M_v$ with the maximum magnitude sum.}
	\label{fig:cl_sample}
\end{figure*}

The evaluation of the proposed curriculum learning strategy is shown in Table \ref{tab:cl_results}. Compared with the baseline results (100\% of K-400), the performances are further boosted on both UCF101 dataset 
and HMDB51 dataset. 
This validates the effectiveness of the proposed curriculum learning strategy. It is also interesting to note that when using the first or the last half of the sorted training samples, \ie, simple samples or difficult samples, the performances on UCF101 dataset are both lower than the random half of K-400. Such observations further validate that the careful selection of training samples is 
necessary
in self-supervised representation learning. Three video samples ranked from easy to hard are shown in Fig. \ref{fig:cl_sample}.

\subsection{\wang{Effectiveness of Training Targets}} \label{sec:target_exp} 

\wang{We evaluate the effectiveness of three different training targets: 1D label regression, 2D label regression, and classification \jb{w.r.t.} three backbone networks in Table~\ref{tab:target}. Specifically, we use UCF101 training split 1 as the pre-training dataset. 
The pretext task performance is evaluated by spatio-temporal statistics prediction accuracy on UCF101 and HMDB51 datasets.
The downstream task performance is evaluated by action recognition accuracy on UCF101 and HMDB51 datasets.}

\wang{We have three key observations: (1) The \jb{proposed} pretext task is quite challenging, when we formulate it as a regression problem. The mean square error loss \jb{introduces ambiguity when predicting discrete numbers}. As a result, it \jb{results in} a poor performance when we \jb{measure the accuracy by considering the exact number}.
(2) \jb{Although} challenging, \jb{the proposed pretext task} indeed encourages the neural networks to learn transferable representation for video understanding. The classification training target achieves the best performances on both pretext task and downstream task \jb{w.r.t.} different network architectures. (3) \jb{A} better pretext task performance \jb{does not always leads to} a better downstream task performance\jb{, in which case too much network capacity may be allocated to the pretext task optimization}. For example, the 2D label regression outperforms the 1D label regression consistently in terms of the pretext task performance. But it achieves worse \jb{downstream (action recognition) task performance in most cases}. }

\begin{table}[]
		\caption{Evaluation of different training targets on UCF101 and HMDB51 datasets w.r.t three backbone networks.}
	    \centering
	    \renewcommand{\arraystretch}{1.1}
	    \large
	    \begin{adjustbox}{max width=\columnwidth}
		\begin{tabular}{c|lcccc}
			\toprule
			~ & Training &
			\multicolumn{2}{c}{Pretext task} &
			\multicolumn{2}{c}{Downstream task} \\
			\cmidrule(lr){3-4}
			\cmidrule(lr){5-6}
			~ & Target  & UCF101 & HMDB51  & UCF101  & HMDB51  \\
			\hline
			\multirow{3}*{\rotatebox{90}{C3D}}& 1D label & 20.5  & 19.1 & 69.3 & 34.2  \\
			~ & 2D label & 24.7 & 23.6 & 68.7 & 30.3 \\
			~ & Classification & \textbf{52.5} & \textbf{49.2} & \textbf{72.3} & \textbf{39.0} \\
			\hline
			\multirow{3}*{\rotatebox{90}{R3D-18}} & 1D label & 18.2 & 17.4 & 67.2  & 32.7  \\
			~ & 2D label & 22.3 & 21.4 & 67.1 & 28.0\\
			~ & Classification & \textbf{49.0} & \textbf{45.5} & \textbf{70.4} & \textbf{34.9} \\
			\hline
			\multirow{3}*{\rotatebox{90}{R(2+1)D}} & 1D label & 19.6 & 18.2 & 73.6 & 34.1  \\
            ~ & 2D label & 23.5 & 22.5 & 74.5 & 32.7 \\
			~ & Classification &  \textbf{50.8} & \textbf{47.6} & \textbf{77.8} & \textbf{40.7} \\
			\bottomrule
		\end{tabular}
		\end{adjustbox}
	\label{tab:target}
\end{table}

\makeatletter
\renewcommand\@seccntformat[1]{\color{black} {\csname the#1\endcsname}\hspace{0.5em}}
\makeatother

\section{Comparison with State-of-the-arts} \label{sec:sota}

In this section, we validate the proposed approach both quantitatively and qualitatively. We compare with state-of-the-arts on four downstream video understanding tasks: action recognition in Section \ref{sec:action}, video retrieval in Section~\ref{sec:retrieval}, dynamic scene recognition in Section~\ref{sec:scene}, and action similarity labeling in Section~\ref{sec:aslan}.

\subsection{Action Recognition} \label{sec:action}

\begin{figure}[thbp!]
	\begin{center}
		\includegraphics[width=\columnwidth]{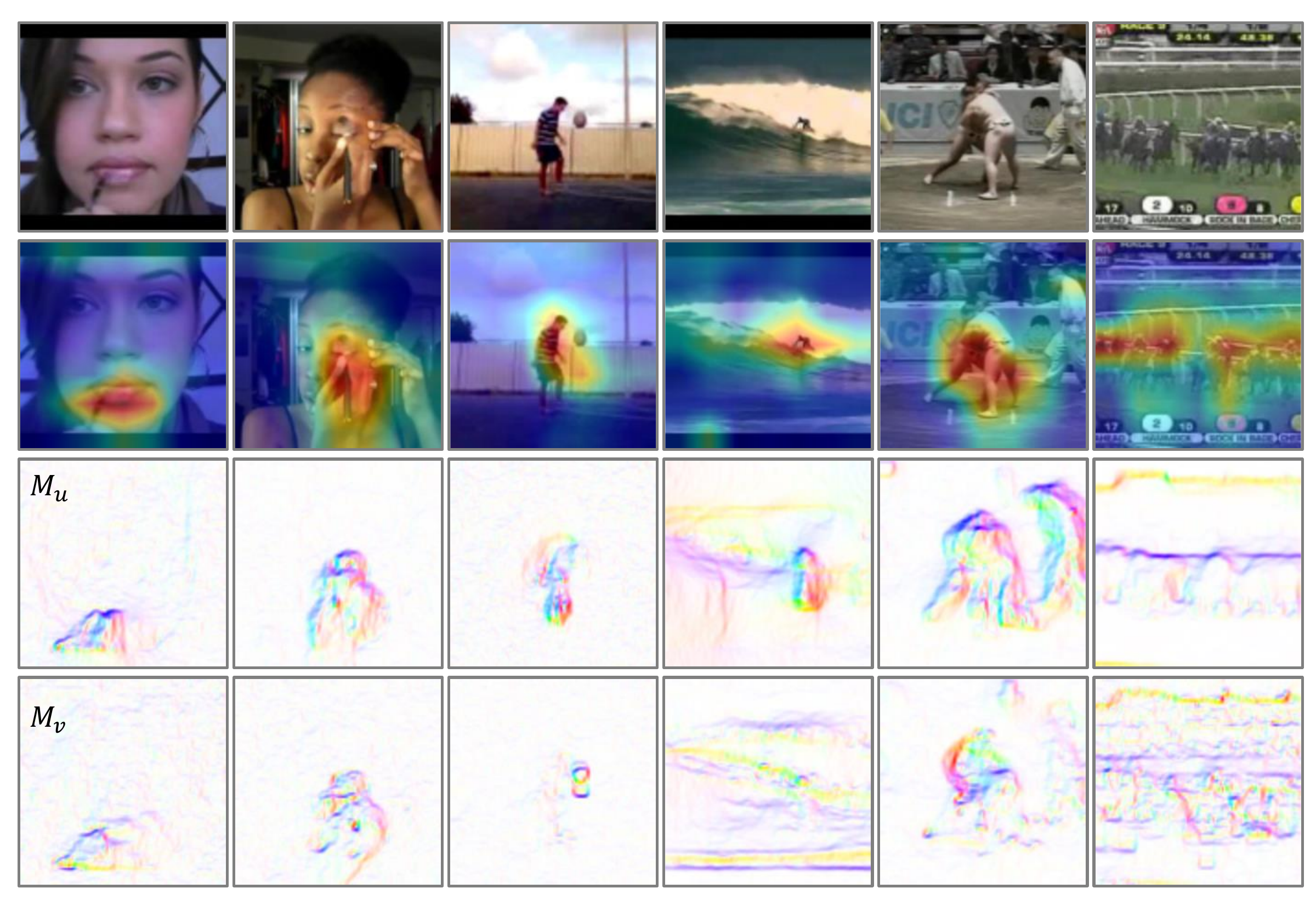} 
	\end{center}
	\caption{Attention visualization. For each sample from top to bottom: A frame from a video clip, activation based attention map of conv5 layer on the frame by using \cite{zagoruyko2016paying}, summarized motion boundaries $M_u$, and summarized motion boundaries $M_v$ computed from the video clip.}
	\label{fig:attention}
\end{figure}

\wang{We compare our approach with state-of-the-art self-supervised learning approaches on the action recognition task in Table \ref{tab:sota}. We have the following observations: (1) Compared with random initialization, \ie, training from scratch, networks fine-tuned on pre-trained models with the proposed spatio-temporal statistics (STS) achieve significant improvement on both UCF101 and HMDB51 datasets. 
Such results demonstrate the great potential of self-supervised video representation learning.
(2) With larger input size, \ie, $112 \times 112$ to $224 \times 224$, longer input length, \ie, 16 frames to 64 frames, and a more powerful backbone network, \ie, R(2+1)D to S3D-G, the performance of the proposed STS can be further improved drastically on both UCF101 and HMDB51 datasets. This \jb{considerably} validates the scalability and potential of the proposed approach.
(3) Our approach achieves state-of-the-art performance on both datasets.
Given RGB videos as the network inputs, the proposed STS \jb{outperforms the state-of-the-art SpeedNet~\cite{benaim2020speednet} by 7.9\% on UCF101 and 13.2\% on HMDB51}.
(4) The proposed approach \jb{even performs better than} MemDPC$^*$~\cite{han2020memory}, which takes both RGB and optical flow as inputs \jb{and is the current best performed method; the proposed approach also achieves comparable performance with AVTS~\cite{korbar2018cooperative} and XDC~\cite{alwassel2020self}}, which use both audio and \jb{video modalities}.
}

\begin{table*}[tbp!]
	\caption{Comparison with state-of-the-art self-supervised learning approaches on the action recognition task. ``$*$'' \jb{indicates} both RGB and optical flow \jb{are taken} as input, where the predictions are finally averaged.}
     \setlength{\tabcolsep}{10pt}
     \begin{center}
     \begin{adjustbox}{max width=2\columnwidth}
		\begin{tabular}{lcccccccc}
			\toprule
			\multicolumn{1}{l}{\multirow{2}{*}{Method}} &
			\multicolumn{6}{c}{Pre-training Experimental Setup} &
			\multicolumn{2}{c}{Downstream task} \\
			\cmidrule(lr){2-7}
			\cmidrule(lr){8-9}
			~ & Network  & Input size &  \#Params & Dataset & Audio & Visual & UCF101 & HMDB51  \\
			\midrule
		    Random & R(2+1)D & $224\times224$ &  14.4M & - && \checkmark &  56.0 & 22.0 \\
		    Fully supervised  & R(2+1)D & $224\times224$ & 14.4M & K-400 && \checkmark & 93.1 & 63.6 \\
			\midrule
			AVTS~\cite{korbar2018cooperative} & I3D & $224\times224$ & 27.2M & K-400 & \checkmark & \checkmark & 83.7 & 53.0 \\
			AVTS~\cite{korbar2018cooperative} & MC3 & $224\times224$ & 11.7M & AudioSet & \checkmark & \checkmark & 89.0 & 61.6 \\
			XDC~\cite{alwassel2020self} & R(2+1)D & $224\times224$ & 14.4M & K-400 & \checkmark & \checkmark & 86.8 & 52.6 \\
			XDC~\cite{alwassel2020self} & R(2+1)D & $224\times224$ & 14.4M & AudioSet & \checkmark & \checkmark & 93.0 & 63.7 \\
			\midrule
			Object Patch~\cite{wang2015unsupervised} & AlexNet  & $ 227 \times 227$ & 62.4M & UCF101 & &\checkmark  & 42.7 & 15.6\\
			ClipOrder~\cite{misra2016shuffle} & CaffeNet  & $ 227 \times 227$ & 58.3M & UCF101 & &\checkmark  & 50.9 & 19.8\\
			AoT~\cite{wei2018learning} &  AlexNet  & $ 227 \times 227$ & 62.4M & UCF101 & &\checkmark  & 55.3 & - \\
			Deep RL~\cite{buchler2018improving} & CaffeNet  & $227 \times 227$ & 58.3M & UCF101& &\checkmark  & 58.6 & 25.0\\
			OPN~\cite{lee2017unsupervised}  & VGG &  $80 \times 80$ & 8.6M & UCF101 &&\checkmark  &59.8 & 23.8 \\
			\midrule
			VCP~\cite{luo2020video} & R(2+1)D & $112\times112$ & 14.4M & UCF101 &&\checkmark  & 66.3 & 32.2\\ 
			VCOP~\cite{xu2019self} & R(2+1)D & $112\times112$ & 14.4M & UCF101 &&\checkmark   & 72.4 & 30.9 \\
			PRP~\cite{yao2020video} & R(2+1)D & $112\times112$ & 14.4M & UCF101 &&\checkmark   & 72.1 & 35.0 \\
			\textbf{STS (Ours)} & R(2+1)D & $112\times112$ & 14.4M & UCF101 && \checkmark & \textbf{77.8} & \textbf{40.7} \\
			\midrule
			MAS~\cite{wang2019self} &  C3D & $112\times112$ & 33.4M & K-400 && \checkmark  & 61.2 & 33.4  \\
			ST-puzzle~\cite{kim2018self} & R3D-18 & $224\times224$ & 33.6M & K-400 && \checkmark&  65.8 & 33.7 \\
			DPC~\cite{han2019video} & R3D-34 & $224\times224$ & 32.6M & K-400 && \checkmark&  75.7 & 35.7 \\
			Pace~\cite{wang2020self} & R(2+1)D & $112\times112$ & 14.4M & K-400 && \checkmark  & 77.1 & 36.6 \\
			MemDPC~\cite{han2020memory} & R-2D3D & $224\times224$ & 32.6M & K-400 && \checkmark  & 78.1 & 41.2 \\TempTrans~\cite{jenni2020video} & R3D-18 & $112\times112$ & 33.6M & K-400 && \checkmark&  79.3 & 49.8 \\
			SpeedNet~\cite{benaim2020speednet} & S3D-G & $224\times224$ & 9.6M  & K-400 &&\checkmark  & 81.1 & 48.8 \\
			MemDPC$^*$~\cite{han2020memory} & R-2D3D & $224\times224$ & 32.6M & K-400 && \checkmark  & 86.1 & 54.5 \\
			\textbf{STS (Ours)} & S3D-G & $224\times224$ & 9.6M & K-400 && \checkmark & \textbf{89.0} & \textbf{62.0} \\
			\bottomrule
		\end{tabular}
		\end{adjustbox}
		\end{center}
	\label{tab:sota}
\end{table*}

\textbf{Attention Visualization.} Fig. \ref{fig:attention} visualizes the attention maps on several video samples using~\cite{zagoruyko2016paying}. For action classes with subtle differences, \eg, \textit{Apply lipstick} and \textit{Apply eye makeup}, the pre-trained model is sensitive to the location that is exactly the largest motion location as quantified by the summarized motion boundaries $M_u$ and $M_v$. It is also interesting to note that for the \emph{SumoWrestling} video sample (the fifth column), although three persons (two players and one judge) have large motion in direction $u$, only players demonstrate larger motion in direction $v$. As a result, the attention map is mostly activated around the players.  

The performances on the action recognition downstream task strongly validate the great power of our self-supervised learning approach. 
The proposed pretext task is demonstrated to be effective in driving backbone networks to learn spatio-temporal features for action recognition. 
In the following, to the goal of learning generic features, we directly evaluate the learned video representation on three different downstream tasks by using the networks as feature extractors without fine-tuning on the downstream task.

\subsection{Video Retrieval} \label{sec:retrieval}
We evaluate spatio-temporal representation learned from our self-supervised approach on video retrieval task. 
Given a video, we follow ~\cite{luo2020video, xu2019self} to uniformly sample ten 16-frame clips.
Then the video clips are fed into the self-supervised pre-trained models to extract features from the last pooling layer (pool5). Based on the extracted video features, cosine distances between videos of testing split and training split are computed. Finally, the video retrieval performance is evaluated on the testing split by querying 
\ang{top-$k$}
nearest neighbours from the training split based on cosine distances. Here, we consider $k$ to be 1, 5, 10, 20, 50. If the test clip class label is within the 
\ang{top-$k$}
retrieval results, it is considered to be successfully retrieved.

\begin{table}[tbp!]
	\caption{Comparison with state-of-the-art self-supervised learning methods on the video retrieval task with the UCF101 dataset. The best results from pool5 w.r.t. each 3D backbone network are shown in \textbf{bold}. The results from pool4 on our method are in \textit{italic} and highlighted.}
	\label{tab:ucf101_fea}
	\small
    \centering
        \begin{adjustbox}{max width=\columnwidth}
		\begin{tabular}{c|lccccc}
			\toprule
			~&Method  & Top-1 & Top-5 & Top-10 & Top-20  & Top-50  \\
			\midrule
			\multirow{3}*{\rotatebox{90}{AlexNet}}&Jigsaw~\cite{noroozi2016unsupervised} & 19.7 & 28.5 & 33.5 & 40.0 & 49.4 \\
			~&OPN~\cite{lee2017unsupervised} & 19.9  & 28.7 & 34.0 & 40.6 & 51.6\\
			~&Deep RL~\cite{buchler2018improving} & 25.7 & 36.2 & 42.2 & 49.2 & 59.5 \\
			\midrule
			\multirow{5}*{\rotatebox{90}{C3D}}&Random& 16.7 & 27.5 & 33.7 & 41.4 & 53.0  \\
			~ & VCOP~\cite{xu2019self} & 12.5 & 29.0 & 39.0 & 50.6 & 66.9\\
			~ & VCP~\cite{luo2020video} & 17.3 & 31.5 & 42.0 & 52.6 & 67.7 \\
			~ & \textbf{Ours} & \textbf{39.1} & \textbf{59.2} & \textbf{68.8} & \textbf{77.6} & \textbf{86.4} \\
		    ~ & \cc{\emph{Ours~(p4)}} & \cc{\emph{43.9}} & \cc{\emph{63.4}} & \cc{\emph{71.3}} & \cc{\emph{79.0}} & \cc{\emph{87.5}}\\
			\midrule
			\multirow{5}*{\rotatebox{90}{R3D-18}}&Random & 9.9 & 18.9 & 26.0 & 35.5 & 51.9  \\
			~&VCOP~\cite{xu2019self} & 14.1 & 30.3 & 40.4 & 51.1 & 66.5\\
			~&VCP~\cite{luo2020video} & 18.6 & 33.6 & 42.5 & 53.5 & 68.1 \\
			~&\textbf{Ours} & \textbf{38.3} & \textbf{59.9} & \textbf{68.9} & \textbf{77.2} & \textbf{87.3} \\
			~ & \cc{\emph{Ours~(p4)}} & \cc{\emph{42.7}} & \cc{\emph{62.3}} & \cc{\emph{71.0}} & \cc{\emph{78.3}} & \cc{\emph{87.3}}\\
			\midrule
			\multirow{5}*{\rotatebox{90}{R(2+1)D}}& Random & 10.6 & 20.7 & 27.4 & 37.4 & 53.1  \\
			~&VCOP~\cite{xu2019self} & 10.7 & 25.9 & 35.4 & 47.3 & 63.9\\
			~&VCP~\cite{luo2020video} & 19.9 & 33.7 & 42.0 & 50.5 & 64.4 \\
			~&\textbf{Ours} & \textbf{38.1} & \textbf{58.9} & \textbf{68.1} & \textbf{77.0} & \textbf{85.9} \\
		    ~ & \cc{\emph{Ours~(p4)}} & \cc{\emph{42.2}} & \cc{\emph{63.4}} & \cc{\emph{71.3}} &\cc{\emph{78.7}} & \cc{\emph{86.9}} \\		
		    \bottomrule
		\end{tabular}
		\end{adjustbox}
\end{table}

\begin{figure*}[htbp!]
	\begin{center}
		\includegraphics[width=\textwidth]{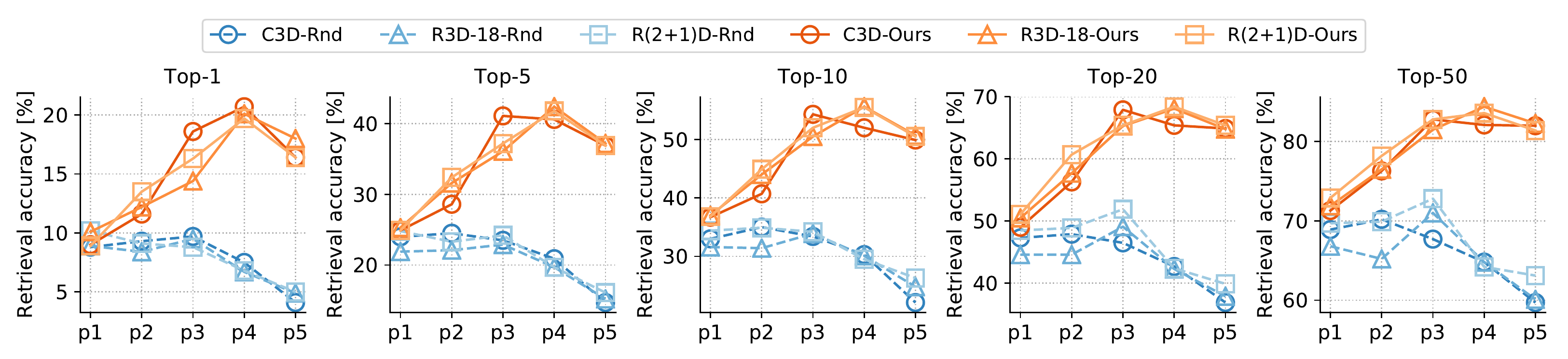} 
	\end{center}
	\vspace{-3mm}
	\caption{Evaluation of features from different stages of the network, \ie, pooling layers, on the video retrieval task with the HMDB51 dataset. The dotted blue lines show the performances of the supervised pre-trained models on the action recognition problem, \ie, random initialization (Rnd). The orange lines show the performances of the self-supervised pre-trained models with our method (Ours). Better visualization with color.}
	\label{fig:vid_retr}
\end{figure*}

In Tables~\ref{tab:ucf101_fea} and~\ref{tab:hmdb51_fea}, we compare our approach with the other self-supervised learning methods on UCF101 dataset and HMDB51 dataset, respectively. The proposed approach 
\ang{outperforms}
VCOP~\cite{xu2019self} and VCP~\cite{luo2020video} on both datasets 
significantly. 
We further investigate whether the performances could be improved, since 
video features extracted from the pool5 layer tend to be more task-specific and could lack generalizability for the video retrieval task. To validate this hypothesis, we extract video features from each pooling layer.
In~Fig. \ref{fig:vid_retr}, we show the comparison between the self-supervised method (pre-trained on the proposed pretext task) and supervised method (pre-trained on the action labels) on HMDB51 dataset, and UCF101 dataset follows the similar 
\ang{trends.}

\begin{figure*}[tbp!]
	\begin{center}
		\includegraphics[width=\textwidth]{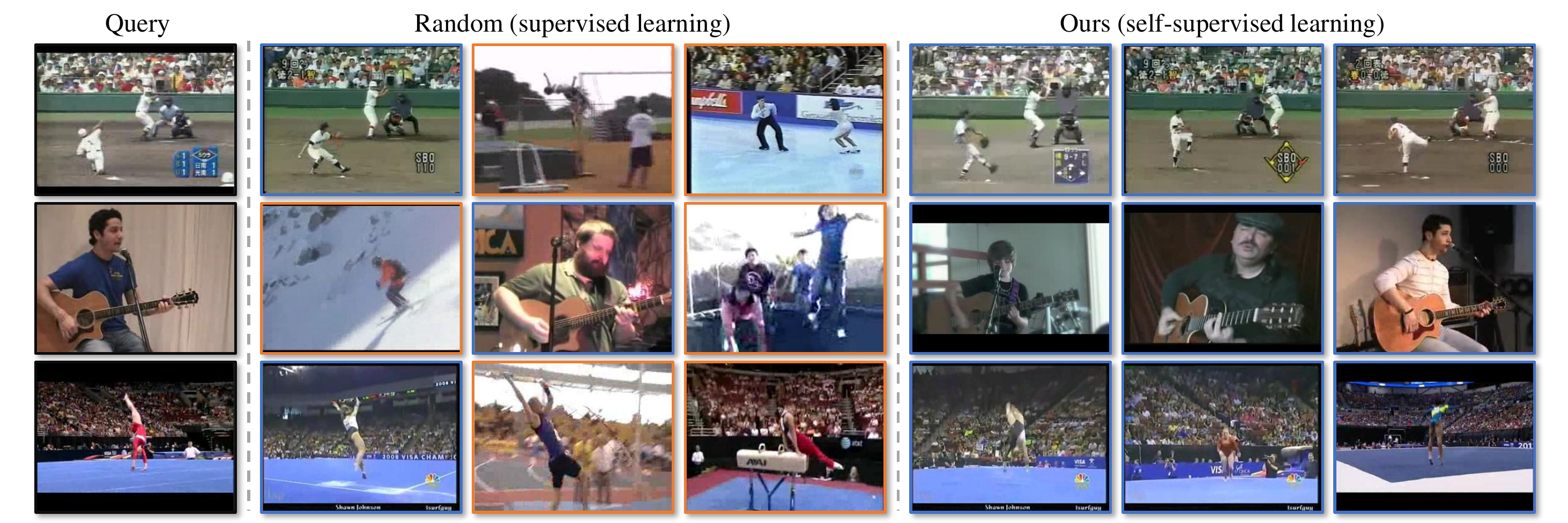} 
	\end{center}
	\vspace{-2mm}
	\caption{Qualitative video retrieval results. From left to right: one query frame from the testing split, frames from the top-3 retrieval results based on the supervised pre-trained models, and frames from the top-3 retrieval results based on our self-supervised pre-trained models. From top to bottom: three qualitative examples of video retrieval on the UCF101 dataset. The correctly retrieved results are marked in blue while the failure cases are in orange. Better visualization with color.}
	\label{fig:vid_retr_sample}
\end{figure*}

We have the following key observations: (1) Regarding our self-supervised method, with the evaluation layer going deeper, the retrieval performance would increase to a peak (usually at pool3 or pool4 layer) and then decrease. Similar observation is also reported in self-supervised image representation learning~\cite{kolesnikov2019revisiting}. The corresponding performance of pool4 layer is reported in Tables~\ref{tab:ucf101_fea} and~\ref{tab:hmdb51_fea} (highlighted in blue). 
(2) Our self-supervised method significantly outperforms the supervised method, especially at deeper layers. This suggests that features learned from our self-supervised method are more robust and generic when transferring to the video retrieval task. Some qualitative video retrieval results are shown in Fig. \ref{fig:vid_retr_sample}. 

\begin{table}[tbp!]
	\caption{Comparison with state-of-the-art self-supervised learning methods on the video retrieval task with the HMDB51 dataset. The best results from pool5 w.r.t. each 3D backbone network are shown in \textbf{bold}. The results from pool4 on our method are in \textit{italic} and highlighted.}
	\label{tab:hmdb51_fea}
	\small
    \centering
        \begin{adjustbox}{max width=\columnwidth}
		\begin{tabular}{c|lccccc}
			\toprule
			&Method& Top-1 & Top-5 & Top-10 & Top-20  & Top-50  \\
			\midrule
			\multirow{5}*{\rotatebox{90}{C3D}}&Random & 7.4 & 20.5 & 31.9 & 44.5 & 66.3  \\
			~& VCOP\cite{xu2019self}& 7.4 & 22.6 & 34.4 & 48.5 & 70.1\\
			~& VCP\cite{luo2020video}& 7.8 & 23.8 & 35.5 & 49.3 & 71.6 \\
			~ & \textbf{Ours} & \textbf{16.4} & \textbf{36.9} & \textbf{49.9} & \textbf{64.9} & \textbf{82.0} \\
			~ & \cc{\emph{Ours~(p4)}} & \cc{\emph{20.7}} & \cc{\emph{36.9}} & \cc{\emph{49.9}} & \cc{\emph{64.9}}  & \cc{\emph{82.0}}\\
			\midrule
			\multirow{5}*{\rotatebox{90}{R3D-18}}&Random & 6.7 & 18.3 & 28.3 & 43.1 & 67.9  \\
			~&VCOP\cite{xu2019self} & 7.6 & 22.9 & 34.4 & 48.8 & 68.9\\
			~&VCP\cite{luo2020video} & 7.6 & 24.4 & 36.6 & 53.6 & 76.4 \\
			~ & \textbf{Ours} & \textbf{18.0} & \textbf{37.2} & \textbf{50.7} & \textbf{64.8} & \textbf{82.3} \\
			~& \cc{\emph{Ours~(p4)}} & \cc{\emph{20.1}} & \cc{\emph{42.4}} & \cc{\emph{55.6}} & \cc{\emph{68.1}} & \cc{\emph{82.3}} \\
			\midrule
			\multirow{5}*{\rotatebox{90}{R(2+1)D}}& Random & 4.5 & 14.8 & 23.4 & 38.9 & 63.0  \\
			~&VCOP\cite{xu2019self} & 5.7 & 19.5 & 30.7 & 45.8 & 67.0\\
			~&VCP\cite{luo2020video} & 6.7 &  21.3 & 32.7 & 49.2 & 73.3\\
			~ & \textbf{Ours} & \textbf{16.4} & \textbf{36.9} & \textbf{50.5} & \textbf{65.4} & \textbf{81.4} \\
			~ & \cc{\emph{Ours~(p4)}} & \cc{\emph{19.7}} & \cc{\emph{41.8}} & \cc{\emph{55.5}} & \cc{\emph{68.4}} & \cc{\emph{83.6}}\\
			\bottomrule
		\end{tabular}
		\end{adjustbox}
\end{table}

\subsection{Dynamic Scene Recognition}\label{sec:scene}

We further study the transferability of the learned features on dynamic scene recognition problem with the YUPENN dataset \cite{derpanis2012dynamic}, which contains 420 video samples of 14 dynamic scenes. Following previous work~\cite{tran2015learning}, each video sample is first split into 16-frame clips with 8 frames overlapped. Then the spatio-temporal feature of each clip is extracted based on the self-supervised pre-trained models from pooling layers. \wang{In practice, similar to Section~\ref{sec:retrieval}, we investigate the best-performing pooing layer w.r.t. each backbone network in such a problem. The best-performing layer for these three networks is \emph{pool4}.}  Next, video-level representation is obtained by averaging the corresponding video-clip features, followed by $L_2$ normalization. Finally, a linear SVM is used for classification and we follow the same leave-one-out evaluation protocol as described in \cite{derpanis2012dynamic}. \wang{We compare our approach with state-of-the-art hand-crafted features and the other self-supervised learning methods in Table \ref{tab:yupenn}. The proposed approach significantly outperforms the state-of-the-art Geometry~\cite{gan2018geometry} by 8.1\%, 6.0\%, and 7.4\% w.r.t. C3D, R3D-18, and R(2+1)D backbone networks, respectively.}

\begin{table}[tbp!]
    \centering
    \small
    \caption{Comparison with state-of-the-art hand-crafted methods and self-supervised representation learning methods on the dynamic scene recognition task.}
    \begin{adjustbox}{max width=\columnwidth}
    \begin{tabular}{lccc}
        \toprule
        Method & Hand-crafted & Self-supervised & YUPENN \\
        \midrule
         SOE~\cite{derpanis2012dynamic}& \checkmark & & 80.7  \\
         SFA~\cite{theriault2013dynamic}& \checkmark & & 85.5  \\
        \midrule
        Object Patch~\cite{wang2015unsupervised}&  &\checkmark & 70.5  \\
        ClipOrder~\cite{misra2016shuffle}&  &\checkmark & 76.7  \\
        Geometry~\cite{gan2018geometry}&  &\checkmark & 86.9  \\
        \midrule
        \textbf{Ours, C3D} & & \checkmark& \textbf{95.0}\\
        Ours, R3D-18 & & \checkmark& 92.9 \\
        Ours, R(2+1)D & & \checkmark& 94.3 \\
        \bottomrule
    \end{tabular}
    \end{adjustbox}
    \label{tab:yupenn}
\end{table}

\subsection{Action Similarity Labeling} \label{sec:aslan}

\begin{table}[htbp!]
    \centering
    \small
    \caption{Comparison with different hand-crafted features and fully-supervised models on the ASLAN dataset.}
    \begin{adjustbox}{max width=\columnwidth}
    \begin{tabular}{lcccc}
        \toprule
        Features & Hand-crafted & Sup. & Self-sup. & Acc. \\
        \midrule
         C3D~\cite{tran2015learning}& & \checkmark & & 78.3 \\
         P3D~\cite{qiu2017learning} & & \checkmark & & 80.8\\
        \midrule
        HOF~\cite{kliper2011action} & \checkmark& & & 56.7 \\
        HNF~\cite{kliper2011action} & \checkmark& & & 59.5 \\
        HOG~\cite{kliper2011action} & \checkmark& & & 59.8 \\
        \midrule
        Ours, C3D&& & \checkmark & 62.0\\
        Ours, R3D-18 & & & \checkmark & 61.7 \\
        \textbf{Ours, R(2+1)D} & & & \checkmark & \textbf{62.1} \\
        \bottomrule
    \end{tabular}
    \end{adjustbox} 
    \label{tab:aslan}
\end{table}

In this section we introduce a challenging downstream task, action similarity labeling. The learned spatio-temporal representation is evaluated on the ASLAN dataset~\cite{kliper2011action}, which contains 3,631 video samples of 432 classes. Unlike action recognition task or dynamic scene recognition task that aims to predict the actual class label, the action similarity labeling task focuses on the \emph{similarity} of two actions.
That is, given two video samples, the goal is to predict whether the two samples are of the same class or not. This task is quite challenging as the test set contains \emph{never-before-seen} actions~\cite{kliper2011action}.   

To evaluate on the action similarity labeling task, we use the self-supervised pre-trained models as feature extractors and use a \textit{linear} SVM for the binary classification, 
\ang{following~\cite{tran2015learning}.}
Specifically, given a pair of videos, each video sample is first split into 16-frame clips with 8 frames overlapped and then fed into the network to extract features from the pool3, pool4 and pool5 layers. The video-level spatio-temporal feature is obtained by averaging the clip features, followed by $L_2$ normalization. After extracting three types of features for each video, we 
compute 12 different distances for each feature as described in \cite{kliper2011action}. The three 12 (dis-)similarities
are concatenated together to obtain a 36-dimensional feature. Since 
\ang{the scales of distances}
are different, we normalize the distances separately into zero-mean and unit-variance, following~\cite{tran2015learning}. A linear SVM is used for classification and we use the 10-fold leave-one-out cross validation same as~\cite{kliper2011action, tran2015learning}.

\wang{We compare our method with full-supervised methods and hand-crafted features in Table~\ref{tab:aslan}. We set a new baseline for the self-supervised method as no previous self-supervised learning methods have been validated on this task. We have the following observations: (1) Our method outperforms the hand-crafted features: HOF, HOG, and HNF (a composition of HOG and HOF). But there is still a big gap between the fully 
supervised method. (2) Unlike the observations in previous experiments (\eg, action recognition), the performances of 
three backbone networks are comparable with each other. We suspect that the reason lies on the fine-tuning scheme leveraged in previous evaluation protocols, where the backbone architecture plays an important role. As a result, we suggest that the proposed evaluation on the ASLAN dataset (Table~\ref{tab:aslan}) could serve as a complementary evaluation task for self-supervised video representation learning to alleviate the influence of backbone networks.}
     
\section{Conclusions}\label{sec:conc}

\wang{In this work, we presented a novel pretext task for self-supervised video representation learning by uncovering a set of spatio-temporal labels derived from motion and appearance statistics. A curriculum learning strategy was incorporated to further improve the representation learning performance. To validate the effectiveness of our approach, we conducted extensive experiments on four downstream tasks of action recognition, video retrieval, dynamic scene recognition, and action similarity labeling, over four different backbone networks, including C3D, R3D-18, R(2+1)D, and S3D-G. Our method achieves
state-of-the-art performances on various \jb{configurations}.
When directly evaluating the learned features by using the pre-trained model as a feature extractor, our approach demonstrates great robustness and transferability to downstream tasks and significantly outperforms the other competing self-supervised methods.
}

\ifCLASSOPTIONcompsoc
\section*{Acknowledgments}
\else
\section*{Acknowledgment}
\fi

This work is partially supported by the Hong Kong RGC TRS under T42-409/18-R, the Hong Kong ITC under Grant ITS/448/16FP, the VC Fund 4930745 of the CUHK T Stone Robotics Institute, the Hong Kong Centre for Logistics Robotics, the Hong Kong-Shenzhen Innovation and Technology Research Institute (Futian), the National Natural Science Foundation of China (No. 61972162), the EPSRC Programme Grant Seebibyte EP/M013774/1, and Visual AI EP/T028572/1.

\ifCLASSOPTIONcaptionsoff
\newpage
\fi



%

\bibliographystyle{IEEEtran}
\bibliography{egbib}




%

\begin{IEEEbiography}
[{\includegraphics[width=1in,height=1.25in,clip,keepaspectratio]{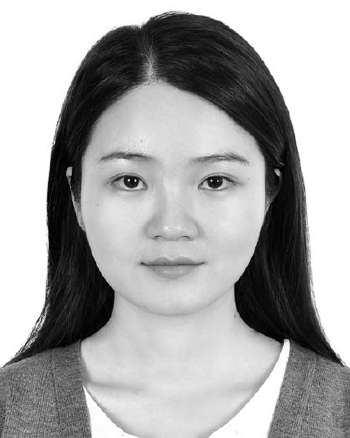}}]
{Jiangliu Wang}
received  the  B.E. degree from Nanjing University in 2015 and the Ph.D. degree from the Chinese University of Hong Kong (CUHK) in 2020, supported by the Hong Kong PhD Fellowship Scheme (HKPFS). She is also co-affiliated with CUHK T Stone Robotics Institute. Her research interests include video understanding, self-supervised representation learning, and related applications in robotics. 
\end{IEEEbiography}

\begin{IEEEbiography}
[{\includegraphics[width=1in,height=1.25in,clip,keepaspectratio]{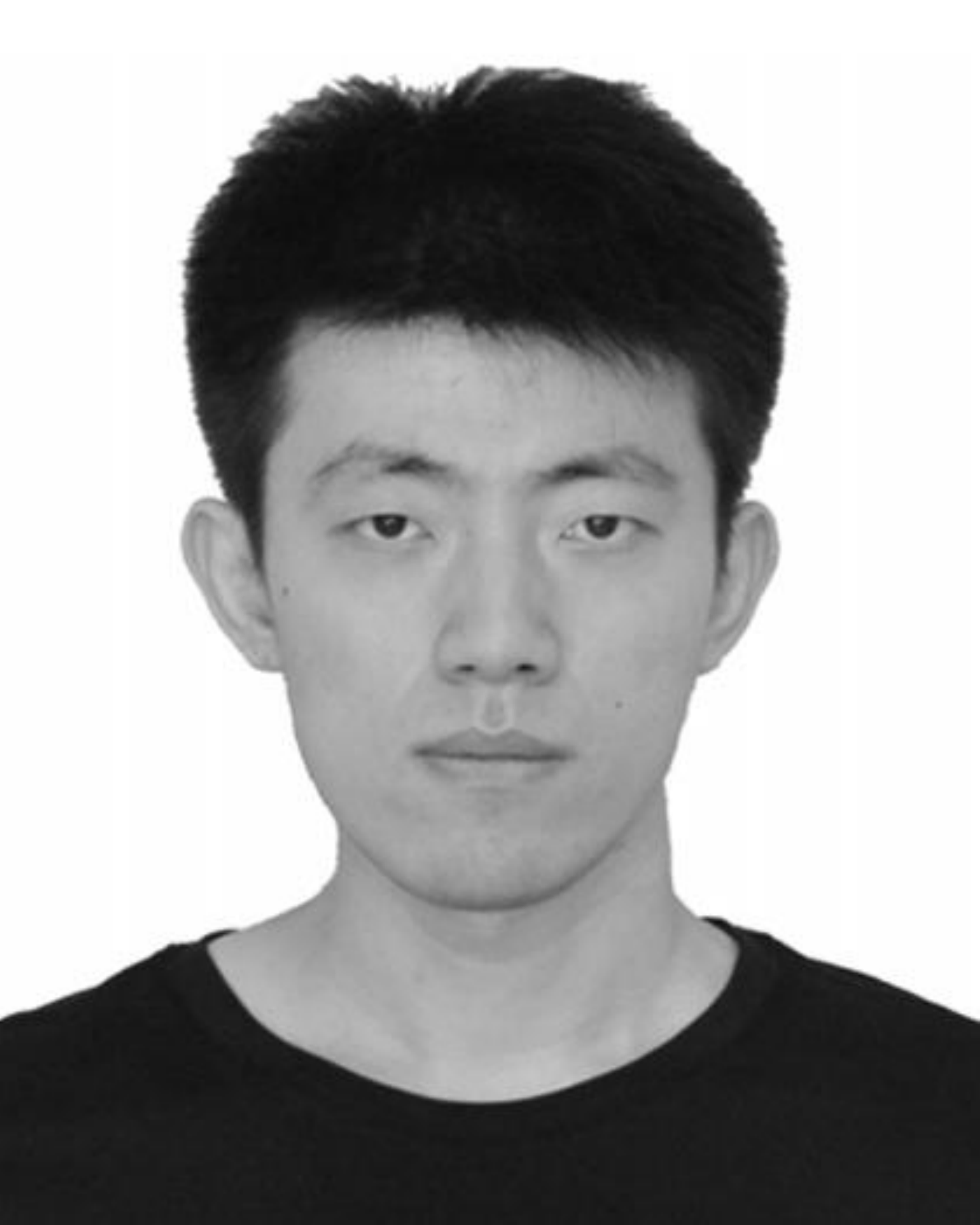}}]
{Jianbo Jiao}(Member, IEEE) received the Ph.D.
degree in computer science from the City University of Hong Kong in 2018, supported by the Hong Kong PhD Fellowship Scheme (HKPFS). He was a Visiting
Scholar with the Beckman Institute, University of
Illinois at Urbana–Champaign from 2017 to 2018.
He is currently a Post-Doctoral Researcher with
the Department of Engineering Science, University
of Oxford. His research interests include computer
vision and machine learning.
\end{IEEEbiography}

\begin{IEEEbiography}
[{\includegraphics[width=1in,height=1.25in,clip,keepaspectratio]{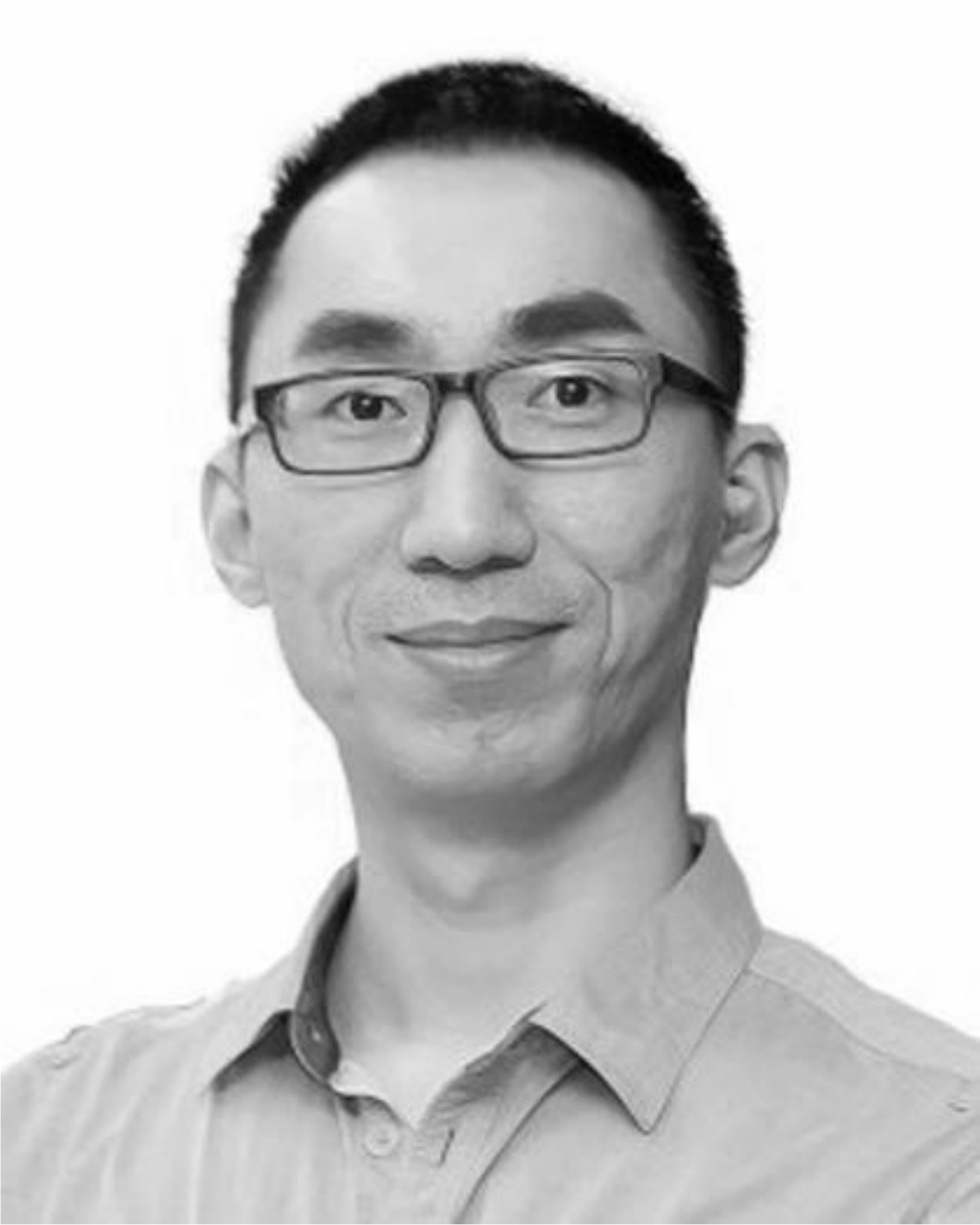}}]
{Linchao Bao} is currently a principal research scientist at Tencent AI Lab. He received the Ph.D. degree in Computer Science from City University of Hong Kong in 2015. Prior to that, he received M.S. degree in Pattern Recognition and Intelligent Systems from Huazhong University of Science and Technology in Wuhan, China. He was a research intern at Adobe Research from November 2013 to August 2014 and worked for DJI as an algorithm engineer from January 2015 to June 2016. His research interests include computer vision and graphics.  
\end{IEEEbiography}

\begin{IEEEbiography}
[{\includegraphics[width=1in,height=1.25in,clip,keepaspectratio]{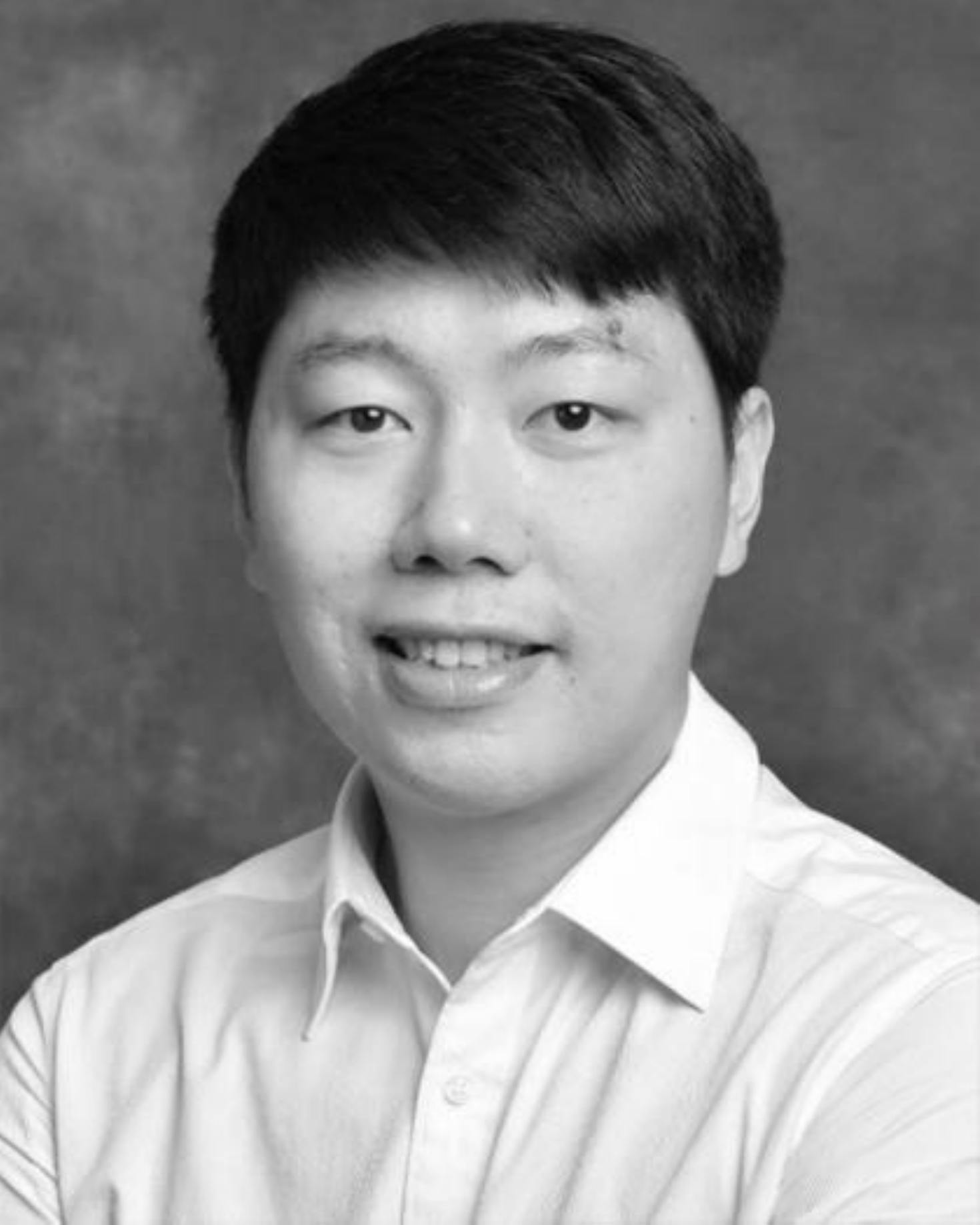}}]
{Shengfeng He} (Senior Member, IEEE) received the B.Sc. and M.Sc. degrees from Macau University of Science and Technology, and a Ph.D. degree from City University of Hong Kong. He is currently an Associate Professor in the School of Computer Science and Engineering, South China University of Technology. His research interests include computer vision, image processing, and computer graphics. He serves as an Associate Editor of the Neurocomputing and IEEE Signal Processing Letter.
\end{IEEEbiography}

\begin{IEEEbiography}
[{\includegraphics[width=1in,height=1.25in,clip,keepaspectratio]{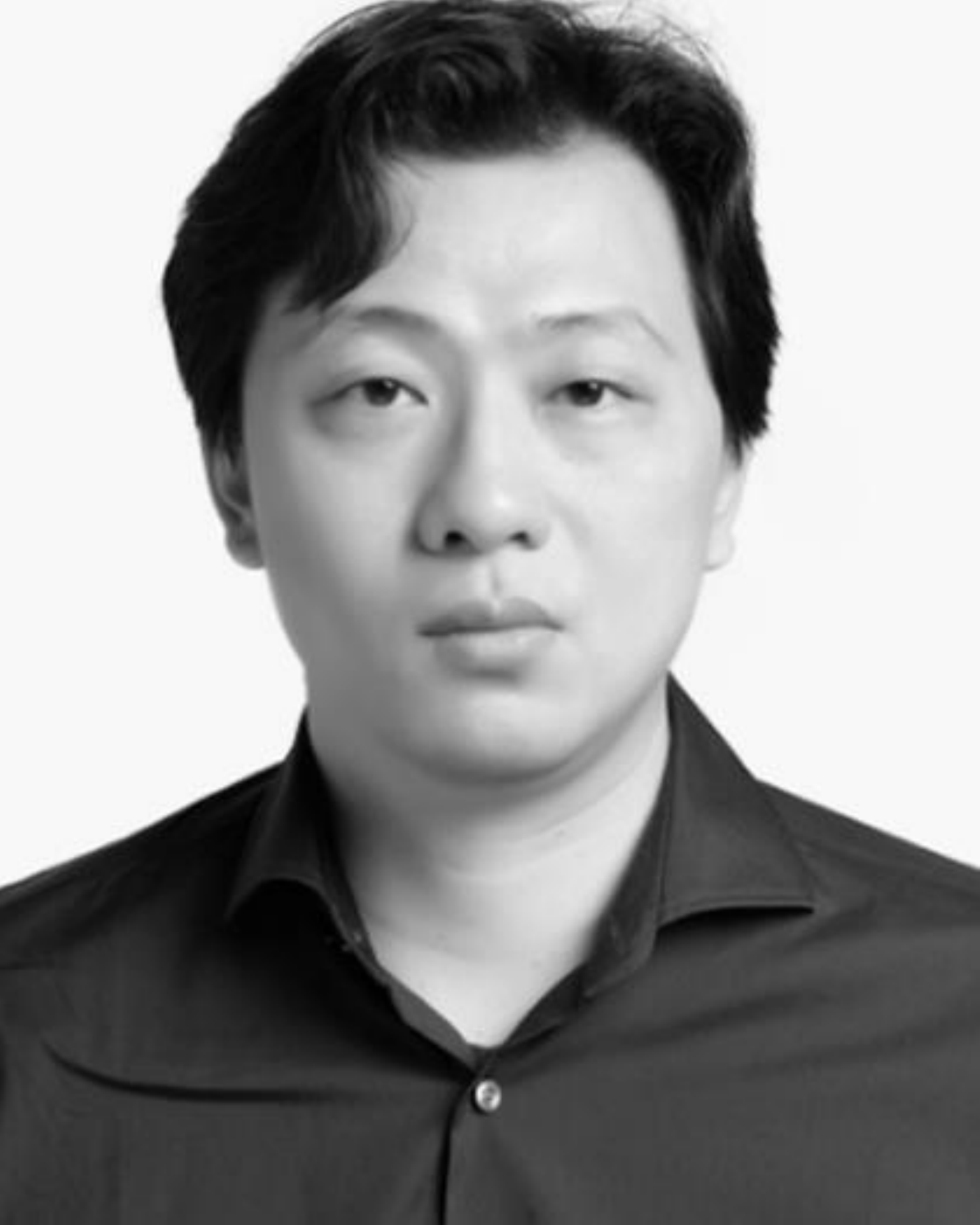}}]
{Wei Liu} (M'14-SM'19) is currently a Distinguished Scientist of Tencent, China and a director of Computer Vision Center at Tencent AI Lab.  Prior to that, he has been a research staff member of IBM T. J. Watson Research Center, Yorktown Heights, NY, USA from 2012 to 2015. Dr. Liu has long been devoted to research and development in the fields of machine learning, computer vision, pattern recognition, information retrieval, big data, etc. Dr. Liu currently serves on the editorial boards of IEEE Transactions on Pattern Analysis and Machine Intelligence, IEEE Transactions on Neural Networks and Learning Systems, IEEE Transactions on Circuits and Systems for Video Technology, Pattern Recognition, etc. He is a Fellow of the International Association for Pattern Recognition (IAPR) and an Elected Member of the International Statistical Institute (ISI). 
\end{IEEEbiography}

\begin{IEEEbiography}
[{\includegraphics[width=1in,height=1.25in,clip,keepaspectratio]{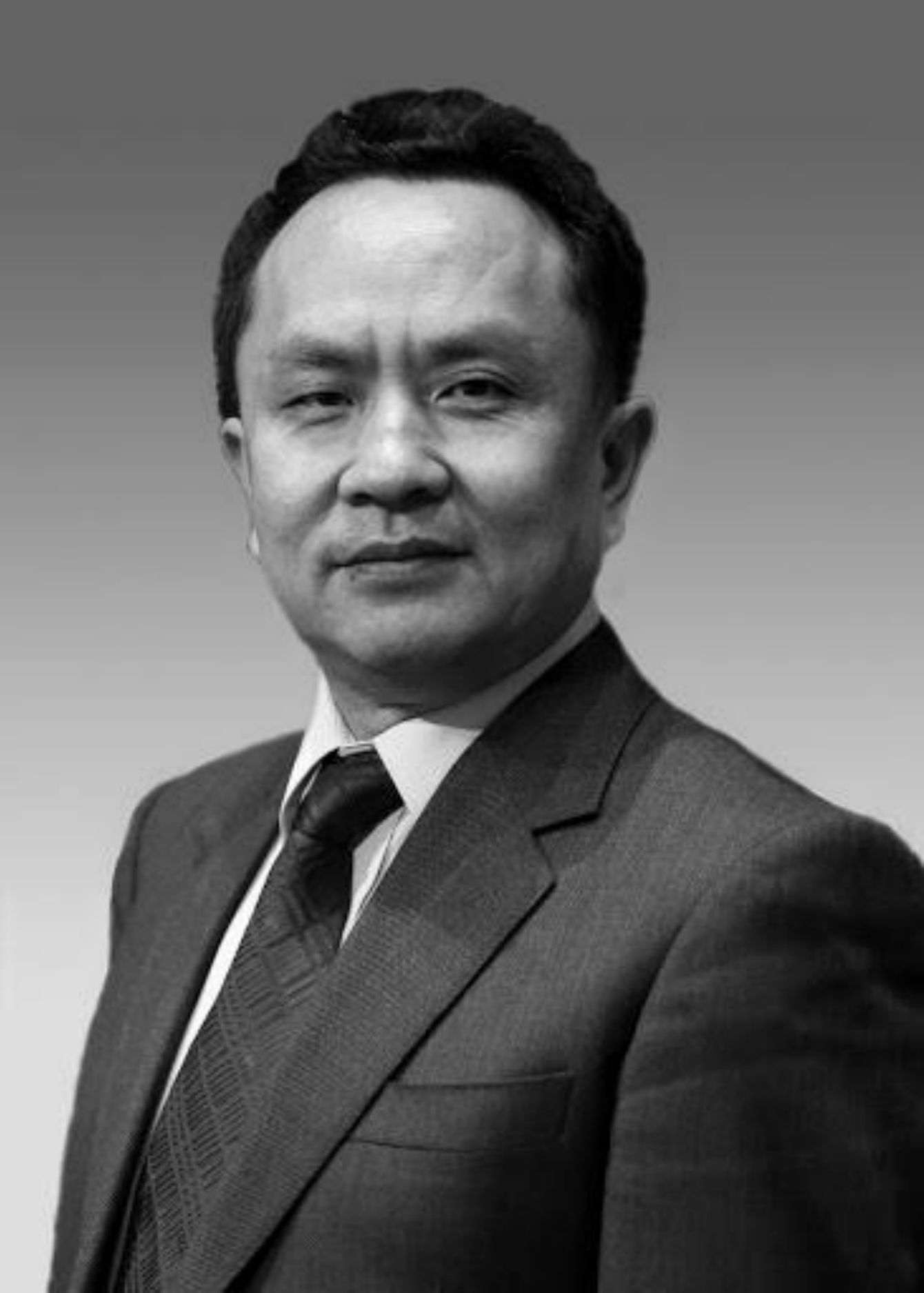}}]
{Yunhui Liu} (Fellow, IEEE) received the B.E. degree in applied
dynamics from the Beijing Institute of Technology,
Beijing, China, the M.E. degree in mechanical
engineering from Osaka University, Suita,
Japan, and the Ph.D. degree in mathematical engineering
and information physics from the University
of Tokyo, Tokyo, Japan, in 1992. He was
with the Electrotechnical Laboratory of Japan,
Tsukuba, Japan, as a Research Scientist, then
he joined The Chinese University of Hong Kong,
Hong Kong, in 1995 and is currently a Choh-Ming Li Professor with the Department of Mechanical and Automation
Engineering, the Director of the CUHK T-Stone Robotics Institute, and
the Director of the Hong Kong Centre for Logistics Robotics.
He has published over 200 papers in refereed journals and refereed
conference proceedings. Dr. Liu was a recipient of the Highly Cited
Author (Engineering) Award by Thomson Reuters in 2013 and numerous
research awards from international journals and international conferences
in robotics and automation and government agencies. He was the
Editor-in-Chief of Robotics and Biomimetics and served as an Associate
Editor of the IEEE TRANSACTIONS ON ROBOTICS AND AUTOMATION, and
General Chair of IROS 2006.
\end{IEEEbiography}





\end{document}